\documentclass{article}

\usepackage{PRIMEarxiv}

\usepackage[utf8]{inputenc}
\usepackage[T1]{fontenc}
\usepackage{hyperref}
\usepackage{url}
\usepackage{booktabs}
\usepackage{amsfonts}
\usepackage{nicefrac}
\usepackage{microtype}
\usepackage{lipsum}
\usepackage{fancyhdr}
\usepackage{graphicx}
\graphicspath{{figures/}}
\usepackage{multirow}
\usepackage[table]{xcolor}
\usepackage{subcaption}
\usepackage{amsmath}
\usepackage{longtable}
\usepackage{natbib}
\usepackage{times}

\title{\textsc{MalruleLib}: Large-Scale Executable Misconception Reasoning with Step Traces for Modeling Student Thinking in Mathematics}

\author{
  Xinghe Chen \\
  Rice University \\
  Houston, TX \\
  \texttt{xc42@rice.edu} \\
  \And
  Naiming Liu \\
  Rice University \\
  Houston, TX \\
  \texttt{nl35@rice.edu} \\
  \And
  Shashank Sonkar \\
  University of Central Florida \\
  Orlando, FL \\
  \texttt{shashank.sonkar@ucf.edu} \\
}

\begin{document}
\maketitle

\begin{abstract}
Student mistakes in mathematics are often systematic: a learner applies a coherent but wrong procedure and repeats it across contexts. We introduce \textsc{MalruleLib}, a learning-science-grounded framework that translates documented misconceptions into executable procedures, drawing on 67 learning-science and mathematics education sources, and generates step-by-step traces of malrule-consistent student work. We formalize a core student-modeling problem as \textbf{Malrule Reasoning Accuracy (MRA)}: infer a misconception from one worked mistake and predict the student’s next answer under cross-template rephrasing. Across nine language models (4B--120B), accuracy drops from 66\% on direct problem solving to 40\% on cross-template misconception prediction. \textsc{MalruleLib} encodes 101 malrules over 498 parameterized problem templates and produces paired dual-path traces for both correct reasoning and malrule-consistent student reasoning. Because malrules are executable and templates are parameterizable, \textsc{MalruleLib} can generate over one million instances, enabling scalable supervision and controlled evaluation. Using \textsc{MalruleLib}, we observe cross-template degradations of 10--21\%, while providing student step traces improves prediction by 3--15\%. We release \textsc{MalruleLib} as infrastructure for educational AI that models student procedures across contexts, enabling diagnosis and feedback that targets the underlying misconception.
\end{abstract}

\section{Introduction}

A student who computes $\frac{1}{2} + \frac{1}{3} = \frac{2}{5}$ is not guessing. They are applying a coherent but flawed procedure. Add the numerators, add the denominators. Learning scientists have shown that many mathematical errors arise from such systematic procedures. They are often called \emph{malrules}, misconceptions, or procedural bugs, and they are stable, diagnosable, and instructionally meaningful \citep{brown1978diagnostic, siegler2013early}. For a tutor, the key step is not verifying correctness. It is inferring \emph{which} malrule a student is using and predicting how it will reappear on the next problem, so feedback targets the underlying reasoning.

Can modern AI do this? \citet{sonkar2025turing} propose a direct ``Educational Turing Test'' for student modeling. Given evidence of a student's misconception, can a model predict the specific errors that student will make on new problems? We operationalize this test with \textbf{Malrule Reasoning Accuracy (MRA)}. In MRA, a model is shown a student's incorrect solution to one problem. The model must infer the underlying malrule from that example and then predict the student's answer on a new problem.

Critically, the new problem often changes surface form. Table~\ref{tab:prompt_comparison} illustrates the challenge. The model sees a student evaluate $\sqrt{x^2 + 25}$ at $x{=}8$ and answer $13$, which implies the malrule $\sqrt{a^2+b^2} = a+b$. The model must then apply the same malrule to a different template, such as a distance word problem that implicitly requires $\sqrt{8^2+3^2}$. This cross-template generalization is what tutoring demands. Students rarely repeat the same question, but they do repeat the same misconception across contexts.

\begin{table}[h]
\centering
\small
\begin{tabular}{lccc}
\toprule
\textbf{Model} & \textbf{CRA} & \textbf{MRA} & \textbf{Forward MRA} \\
\midrule
Llama-3.3-70B & 70.4\% & 34.6\% & 39.6\% \\
Qwen3-80B-Think & 70.1\% & 56.4\% & 53.9\% \\
gpt-oss-120b & 65.0\% & 56.9\% & 48.8\% \\
\midrule
\textit{Average} & 68.5\% & 49.3\% & 47.5\% \\
\bottomrule
\end{tabular}
\caption{Three evaluation settings for the educational Turing Test. \textbf{CRA}: solve problems correctly. \textbf{MRA}: infer misconceptions from examples and predict student answers. \textbf{Forward MRA}: receive explicit misconception descriptions and predict answers.}
\label{tab:intro_gap}
\end{table}

\begin{table*}[t]
\centering
\small
\begin{tabular}{p{0.47\textwidth}|p{0.47\textwidth}}
\toprule
\textbf{Forward MRA} & \textbf{MRA (Cross-Template)} \\
\midrule
\textit{System:} You are simulating a student who has a specific mathematical misconception. Apply the described misconception consistently to solve the problem.
&
\textit{System:} You are an expert in identifying and understanding student mathematical misconceptions. Given an example of a student's incorrect answer, identify the systematic error and apply it to predict answers for new problems.
\\[1em]
\textit{User:} A student has the following misconception:

\textbf{Students distribute square root over addition: $\sqrt{a^2+b^2} = a + b$}

Apply this misconception to solve:

Evaluate $f(x) = \sqrt{x^2 + 4}$ when $x = 3$. What is $f(3)$?
&
\textit{User:} A student solved this problem incorrectly:

\textbf{Problem:} Evaluate $f(x) = \sqrt{x^2 + 25}$ when $x = 8$.

\textbf{Student's Answer:} 13

Now predict what this same student would answer for:

You walk 8 blocks east and 3 blocks north. What is the straight-line distance from your starting point?
\\
\midrule
\textit{Expected:} 5 \hfill (correct: $\sqrt{13} \approx 3.61$)
&
\textit{Expected:} 11 \hfill (correct: $\sqrt{73} \approx 8.54$)
\\
\bottomrule
\end{tabular}
\caption{Prompts for Forward MRA and MRA tasks. \textbf{Forward MRA} provides an explicit misconception description; the model must translate it into procedural errors. \textbf{MRA} provides only a worked example; the model must infer the misconception pattern and generalize it to a new problem format (here, from algebraic to word problem). Both prompts target the same underlying malrule: distributing square roots over addition.}
\label{tab:prompt_comparison}
\end{table*}

\begin{table*}[t]
\centering
\small
\begin{tabular}{p{1.8cm}p{2.5cm}p{5.5cm}p{4.5cm}}
\toprule
\textbf{Category} & \textbf{Malrule} & \textbf{Problem} & \textbf{Student's Work} \\
\midrule
\multirow{2}{*}{Radicals} & \multirow{2}{*}{$\sqrt{a^2+b^2}=a+b$}
& \textbf{T1:} Evaluate the function $f(x) = \sqrt{x^2 + 25}$ when $x = 8$. What is $f(8)$?
& $\sqrt{x^2} + \sqrt{25} = x + 5$; $8 + 5 = \mathbf{13}$ \\
& & \textbf{T2:} You walk 8 blocks east and 3 blocks north. What is the straight-line distance from your starting point?
& $\sqrt{8^2} + \sqrt{3^2} = 8 + 3 = \mathbf{11}$ \\
\midrule
\multirow{2}{*}{\shortstack[l]{Order of\\Operations}} & \multirow{2}{*}{\shortstack[l]{Addition before\\subtraction}}
& \textbf{T1:} Evaluate: $29 - 28 + 12$
& $28 + 12 = 40$; $29 - 40 = \mathbf{-11}$ \\
& & \textbf{T2:} Starting at 45°F, the temperature decreases by 5°F, then increases by 3°F. What's the result?
& $5 + 3 = 8$; $45 - 8 = \mathbf{37}$°F \\
\midrule
\multirow{2}{*}{Functions} & \multirow{2}{*}{\shortstack[l]{$f(a+b)=$\\$f(a)+f(b)$}}
& \textbf{T1:} Given $f(x) = x^3$, evaluate $f(11+10)$
& $f(11)=1331$, $f(10)=1000$; $1331+1000 = \mathbf{2331}$ \\
& & \textbf{T2:} Given $f(x) = |x+3|$, evaluate $f(8+4)$
& $f(8)=11$, $f(4)=7$; $11+7 = \mathbf{18}$ \\
\midrule
\multirow{2}{*}{Division} & \multirow{2}{*}{\shortstack[l]{Larger $\div$ smaller\\always}}
& \textbf{T1:} 4 cookies are shared equally among 6 children. How much does each child get?
& $4 \div 6 \rightarrow 6 \div 4 = \mathbf{1.5}$ \\
& & \textbf{T2:} 4 meters of ribbon is divided into 5 equal strips. How long is each strip in meters?
& $4 \div 5 \rightarrow 5 \div 4 = \mathbf{1.25}$ \\
\midrule
\multirow{2}{*}{Subtraction} & \multirow{2}{*}{\shortstack[l]{Borrow without\\decrementing}}
& \textbf{T1:} Calculate: $408 - 384$
& $8{-}4{=}4$, $10{-}8{=}2$, $4{-}3{=}1$ $\rightarrow$ $\mathbf{124}$ \\
& & \textbf{T2:} A store had 561 items in stock. After selling 526 items, how many remain?
& $11{-}6{=}5$, $6{-}2{=}4$, $5{-}5{=}0$ $\rightarrow$ $\mathbf{45}$ \\
\bottomrule
\end{tabular}
\caption{Examples of malrules from \textsc{MalruleLib}, showing two templates per misconception. Each malrule produces systematic errors across different problem formats: algebraic expressions and word problems. The template diversity illustrates the cross-template generalization challenge: models must recognize the same underlying misconception despite surface-level differences. See Appendix Table~\ref{tab:malrule_ex_examples} for additional examples.}
\label{tab:malrule_examples}
\end{table*}

Our results expose a sharp gap between doing mathematics and modeling student thinking. On three representative large models (70B--120B) shown in Table~\ref{tab:intro_gap}, models achieve 68.5\% accuracy on direct problem solving (\textbf{CRA}: Correct Reasoning Accuracy), but only 49.3\% on \textbf{cross-template} misconception prediction from an example (\textbf{MRA}). \textbf{Mathematical reasoning ability does not transfer to student modeling.} We also evaluate \textbf{Forward MRA}, where the model is given an explicit natural-language description of the misconception and must apply it, and find performance remains limited (47.5\% on the same models). Table~\ref{tab:consolidated} reports the full benchmark across all nine models and experimental settings.

Misconceptions are well studied, but the field lacks infrastructure that treats them as computational objects. Most resources describe misconceptions, but they do not operationalize them as procedures that can be executed across many templates with malrule-consistent intermediate steps \citep{lucy2024evaluating}. This makes it difficult to generate training data, run controlled evaluations, or measure cross-template student modeling at scale.

We address this gap with \textsc{MalruleLib}\footnote{Code and models are available at \url{https://github.com/luffycodes/malrulelib}}, a learning-science-grounded framework that encodes misconceptions as executable procedures and pairs them with diverse problem templates. For each instantiated problem, \textsc{MalruleLib} generates dual-path solution traces: a fully correct solution and a malrule-consistent student solution, both with step-by-step work. Because malrules are executable and templates are parameterizable, \textsc{MalruleLib} can generate large-scale data with ground-truth malrule identity and trace-level supervision. Table~\ref{tab:malrule_examples} illustrates the structure. Each malrule is shown on two different templates, often shifting from a symbolic expression to a word problem. The student's work remains systematically consistent with the same underlying procedure, even as surface features change. This is the core challenge for personalized learning and for our benchmark: models must infer the malrule from evidence and predict misconception-consistent work under cross-template shifts, not merely reproduce a template-specific error pattern.

We make three contributions:
\begin{enumerate}
    \item \textbf{A learning-science-grounded misconception library as executable procedures.} We translate 101 documented malrules into computational objects: each misconception is implemented as an executable procedure and paired with 498 diverse problem templates spanning 22 mathematical categories. Each malrule is grounded in the learning-science literature, drawing from 67 papers (Appendix Tables~\ref{tab:malrulesoc}--\ref{tab:malrulesoc4}). This grounding ensures the benchmark targets real, instructionally meaningful error patterns and supports downstream tutoring actions beyond prediction.

    \item \textbf{Dual-path solution traces at scale.} For every template instance, we generate aligned step-by-step work for \textit{both} correct reasoning and malrule-consistent student reasoning. Because templates are parameterizable and malrules are executable, \textsc{MalruleLib} can generate millions of such paired instances with dual traces, providing trace-level supervision for training and controlled evidence for evaluation. In our experiments, supplying student steps yields substantial improvements in misconception prediction, validating the value of step-level data for student modeling.

    \item \textbf{The first large-scale benchmark for cross-template misconception prediction.} We evaluate 9 language models (4B--120B) on \textbf{Malrule Reasoning Accuracy} under controlled conditions that separate same-template from cross-template generalization and compare answer-only versus with-steps evidence. Across models, cross-template performance drops by 10--21 points, and step evidence produces consistent gains of 3--15 points, exposing a persistent gap between problem solving and modeling misconception-driven student behavior.
\end{enumerate}

\section{Related Work}

\begin{quote}
\itshape
If you can both listen to children and accept their answers not as things to just be judged right or wrong but as pieces of information which may reveal what the child is thinking you will have taken a giant step toward becoming a master teacher rather than merely a disseminator of information.
\end{quote}
\noindent\hfill\citet{easley1975teaching}

\noindent This ``teaching by listening'' view motivates a diagnostic stance toward student errors. The goal is not only to mark an answer right or wrong, but to infer the underlying procedure that produced it. \citet{brown1978diagnostic} operationalized this idea in the \textsc{BUGGY} line of work. They modeled systematic errors as small, structured edits to a correct procedure, and used these diagnostic models to explain errors, predict future mistakes, and even generate diagnostic tests. \textsc{MalruleLib} follows this tradition and extends it to a broader range of mathematical domains and to modern evaluation of language models.

\subsection{The \textsc{BUGGY} Tradition: Procedural Misconceptions in Learning Science}

The study of mathematical misconceptions has deep roots in cognitive science, but \textsc{BUGGY} marked an important shift in emphasis. Rather than treating wrong answers as noise, \citet{brown1978diagnostic} argued that many errors are coherent, rule-governed procedures. In their framing, a child writing $\frac{1}{2} + \frac{1}{3} = \frac{2}{5}$ is not confused about addition. The child is overgeneralizing whole-number procedures to fractions. This procedural view transformed how educators and tutoring systems reason about student knowledge. The target of instruction becomes the underlying rule, not the surface error.

Subsequent decades uncovered misconceptions across mathematical domains. \citet{behr1984order} and \citet{ni2005teaching} documented ``whole number bias'' in fraction arithmetic, where students treat $\frac{a}{b}$ as two independent numbers rather than a single quantity. \citet{resnick1989conceptual} identified systematic errors in decimal comparison, such as believing $0.29 > 0.3$ because 29 is greater than 3. \citet{matz1980towards} catalogued algebraic misconceptions, and \citet{vlassis2004making} traced difficulties with signed numbers to overgeneralized subtraction rules.

A key finding across this literature is that misconceptions are remarkably stable. \citet{siegler2013early} show that fraction magnitude understanding in middle school predicts mathematical achievement years later. Once formed, faulty mental models resist correction. This stability makes misconception prediction both tractable and educationally valuable. \textsc{MalruleLib} operationalizes this literature as executable procedures. It encodes malrules along with prevalence information, root-cause hypotheses, and remediation strategies, then uses them to generate structured student work.

\subsection{Misconception Datasets and Benchmarks}

Prior work has developed resources for studying student errors, but existing datasets share common limitations. ASSISTments \citep{heffernan2014assistments} logs student interactions at scale but it does not capture actual student solutions.
More recent work has begun addressing the rationale gap. \citet{sonkar2024malalgoqa} introduce \textsc{MalAlgoQA}, a dataset of roughly 807 mathematics comprehension questions, each annotated with misconceptions. They find that LLMs exhibit substantial drops when identifying misconceptions compared to correct-answer rationales, foreshadowing our CRA--MRA gap. However, MalAlgoQA provides static rationales curated from existing items rather than executable misconception procedures grounded in learning-science research. It also evaluates questions largely in isolation, without controlled cross-template generalization tests of the same misconception across surface forms.

\subsection{AI for Education and Student Modeling}

Student modeling is central to intelligent tutoring systems, such as cognitive tutors \citep{anderson1995cognitive}, LLM based tutors \citep{sonkar2023class,sonkar2024pedagogical} and knowledge tracing \citep{corbett1994knowledge,sonkar2020qdkt,sonkar2023deduction}. These approaches work well in narrow domains but require substantial hand engineering.
\citet{sonkar2025turing} argue that educational AI should be evaluated by whether it can predict student behavior, a Turing-like test for personalized education.  Our work provides the first large-scale benchmark for this capability.
In doing so, we connect modern LLM evaluation to the earlier diagnostic modeling tradition exemplified by \textsc{BUGGY}.

\section{The MalruleLib Framework}

\begin{figure*}[t]
\centering
\begin{minipage}{0.44\textwidth}
\centering
\small
\setlength{\tabcolsep}{2pt}
\begin{tabular}{@{}lccp{2.9cm}@{}}
\toprule
\textbf{Category} & \textbf{M} & \textbf{T} & \textbf{Description} \\
\midrule
\multicolumn{4}{l}{\textsc{Number \& Operations} (54 malrules, 277 templates)} \\
Whole Number Ops & 16 & 97 & Place value, regrouping \\
Fractions \& Ratios & 13 & 63 & Part-whole, proportions \\
Decimals \& Percents & 16 & 83 & Notation, conversions \\
Signed Numbers & 9 & 34 & Integers, absolute value \\
\midrule
\multicolumn{4}{l}{\textsc{Algebra} (37 malrules, 168 templates)} \\
Exponents \& Radicals & 12 & 72 & Laws of exponents, roots \\
Expressions \& Equations & 21 & 84 & Simplifying, PEMDAS \\
Functions & 4 & 12 & Notation, input-output \\
\midrule
\multicolumn{4}{l}{\textsc{Geometry \& Measurement} (8 malrules, 20 templates)} \\
Geometry & 6 & 14 & Area, perimeter, volume \\
Coordinate Geometry & 2 & 6 & Ordered pairs, graphing \\
\midrule
\multicolumn{4}{l}{\textsc{Data \& Modeling} (4 malrules, 33 templates)} \\
Data \& Word Problems & 4 & 33 & Statistics, translation \\
\bottomrule
\end{tabular}
\label{tab:nctm_classification}
\end{minipage}
\hfill
\begin{minipage}{0.55\textwidth}
\centering
\includegraphics[width=\textwidth]{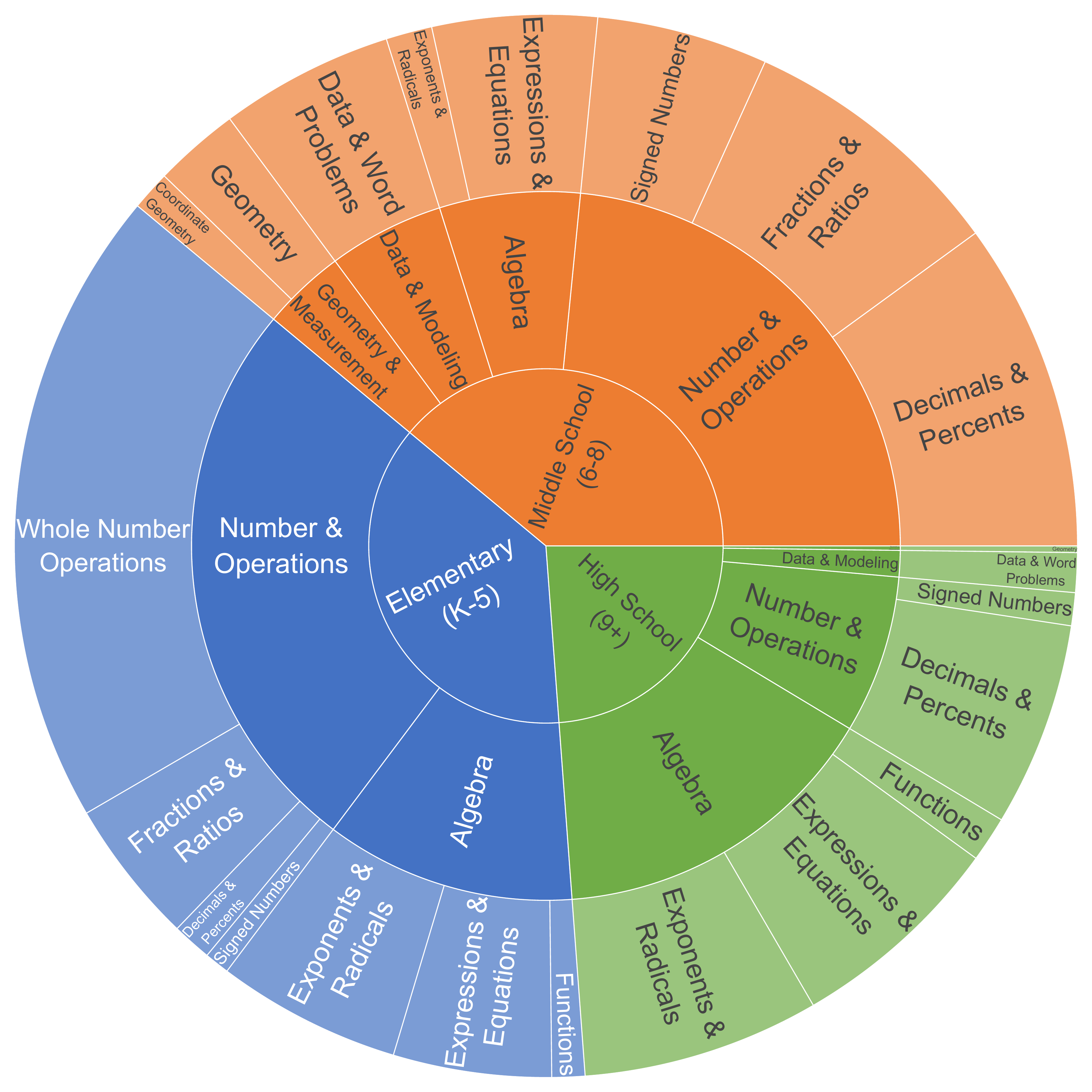}
\label{fig:sunburst}
\end{minipage}
\caption{MalruleLib framework overview. \textbf{Left}: Classification by NCTM Content Strand (M=Malrules, T=Templates). \textbf{Right}: Distribution by developmental stage showing coverage across grade levels.}
\label{fig:framework_overview}
\end{figure*}

Building AI that understands student thinking requires data that captures how students actually reason when making mistakes. We need not just wrong answers, but the cognitive processes behind them. \textsc{MalruleLib} is a Python framework designed to generate such data at scale, grounded in decades of learning science research (see Table~\ref{tab:malrulesoc}).

\subsection{Design Principles}

Three principles guided our framework design.

First, \textbf{learning science grounding}. Every malrule in \textsc{MalruleLib} traces to documented research on student misconceptions. We don't invent plausible errors; we encode errors that real students consistently make, complete with prevalence data and cognitive explanations. This grounds the framework in empirical findings rather than speculation about what might confuse students. A full mapping of malrules to their academic sources appears in Appendix Tables~\ref{tab:malrulesoc}--\ref{tab:malrulesoc4}.

Second, \textbf{cognitively faithful student solutions}. A critical design decision: for every problem, we generate step-by-step solutions showing \textit{both} correct and incorrect reasoning paths. The malrule path captures how students actually think when applying a misconception---not just wrong answers, but the cognitive process behind them. This dual-path generation enables training LLMs to understand and predict student reasoning, not merely evaluate correctness.

Third, \textbf{template diversity}. Each malrule is instantiated through multiple templates: basic formulations, structural variants, real-world contexts, and word problems. The 498 templates (4.9 per malrule on average) enable rigorous cross-template generalization testing---assessing whether models truly understand misconceptions or merely pattern-match surface features.

\paragraph{Malrule Architecture} \textsc{MalruleLib} is open-source and designed for extensibility. Each malrule is a self-contained module with four components:

\small
\begin{verbatim}
[category]/[malrule_name]/
  |--problem_generator.py  # Templates
  |--correct_algorithm.py  # Correct steps
  |--malrule_algorithm.py  # Malrule steps
  |--test_malrule.py       # Unit tests
\end{verbatim}
\normalsize

The \textbf{problem generator} defines multiple templates---parameterized problem structures that produce diverse instances. The \textbf{correct algorithm} and \textbf{malrule algorithm} implement the mathematically sound and misconception-based procedures respectively, each generating step-by-step reasoning.
This modular design makes it straightforward to add new malrules: implement the four components, and the framework handles problem generation, validation, and integration.

\subsection{Coverage Statistics}

\textsc{MalruleLib} currently encodes \textbf{101 malrules} spanning common student errors from elementary through early high school mathematics, organized into \textbf{22 mathematical categories} aligned with 10 NCTM content strands (Figure~\ref{fig:framework_overview}). These malrules are instantiated through \textbf{498 templates} (4.9 per malrule on average), enabling generation of thousands of unique problem instances with \textbf{dual-path step generation} for every problem. Categories range from elementary arithmetic (whole number operations, basic fractions) through middle school topics (exponents, linear equations) to early algebra (factoring, functions). Table~\ref{tab:category_breakdown} shows the distribution across NCTM strands.

Deep analysis of these 498 templates reveals intentional pedagogical design along two dimensions grounded in learning science.

\textbf{Context domain diversity.} Templates span 10 context domains: 63.3\% use abstract mathematical notation while 36.7\% embed problems in real-world contexts such as measurement (11.0\%), money (9.2\%), time/distance (3.8\%), science (3.0\%), sports (3.0\%), and food (3.0\%).
This design reflects research on \textit{transfer of learning}: students often fail to apply knowledge across contexts, exhibiting ``inert knowledge'' that activates only in familiar settings \citep{bransford1999people}. A student may add fractions correctly in abstract form but apply the malrule when the same problem appears as pizza slices. By varying context while holding the misconception constant, we assess whether models exhibit similar context-dependent failures.

\textbf{Scaffolded complexity.} Templates follow a pedagogical progression: \textit{basic} (18.5\%), \textit{variant} (50.8\%), \textit{context} (6.2\%), and \textit{word problem} (24.5\%). This scaffolding enables assessment across difficulty gradients. Word problems particularly test whether misconceptions persist under additional cognitive load from reading comprehension. Additional template coverage details appear in the Appendix \ref{appendix:template}.

\subsection{A Misconception Generator at Million-Instance Scale}

\textsc{MalruleLib} is not just a collection of misconception labels. It is a \emph{generator} that can produce over one million distinct problem instances with different parameters and surface forms. The core idea is simple: if misconceptions are stable procedures, they should be encoded as \emph{executable programs}. Once a malrule is executable, we can systematically generate both (i) the correct solution trace and (ii) the misconception-consistent student trace, at scale and under tight control.

Templates are parameterized by grade-banded value ranges and difficulty presets, enabling both systematic coverage and pedagogically meaningful distributions. We estimate generation capacity by counting valid parameter assignments and template variants that satisfy the constraints for a chosen grade and difficulty setting yielding over one million distinct instances. This scale and mechanism makes it feasible to fine-tune or instruction-tune models directly on executed misconception traces. Additionally, each template exposes a small set of parameters together with malrule-specific constraints that guarantee the misconception is actually triggered. For borrowing-related subtraction, for example, \textsc{MalruleLib} constructs operands digit by digit and enforces the inequalities that force borrowing at the intended place value.

\section{Benchmark Design}

Using \textsc{MalruleLib}, we construct a benchmark that operationalizes the Turing Test for educational AI: can models predict student errors well enough to demonstrate understanding of student reasoning?

\subsection{Task Definitions}

We introduce notation to formally define our evaluation tasks. Let $m \in \mathcal{M}$ denote a malrule from set of 101 malrules. Each malrule has an associated set of templates $\mathcal{T}_m = \{t_1, t_2, \ldots\}$. An instance $i \sim t$ is a concrete problem sampled from template $t$ by instantiating its parameters with specific values. For each instance $i$, we denote $a_c(i)$ and $a_m(i)$ as the correct and malrule answers respectively, and $S_m(i)$ as the step-by-step malrule reasoning.

\paragraph{Malrule Reasoning Task.} Given a source instance $i_s$ with its malrule answer $a_m(i_s)$, and a target problem $i_t$, the model must predict $a_m(i_t)$, the answer this student would give if applying the same malrule. We vary two factors: \textit{template condition} and \textit{prompt condition}. For template condition, \textit{same-template} samples source and target from the same template ($i_s, i_t \sim t$), while \textit{cross-template} samples from different templates ($i_s \sim t_1, i_t \sim t_2$ where $t_1 \neq t_2$). For prompt condition, \textit{answer-only} provides the source problem and malrule answer, while \textit{with-steps} additionally provides the malrule reasoning steps $S_m(i_s)$. Table~\ref{tab:mra_conditions} summarizes the four experimental conditions.

\begin{table}[h]
\centering
\setlength{\tabcolsep}{4pt}
\begin{tabular}{lc}
\toprule
\textbf{Condition} & \textbf{Predict} \\
\midrule
\multicolumn{2}{l}{\textit{Same-template}: $i_s, i_t \sim t$} \\
\quad Answer-only: $(i_s, a_m(i_s), i_t)$ & $a_m(i_t)$ \\
\quad With-steps: $(i_s, a_m(i_s), S_m(i_s), i_t)$ & $a_m(i_t)$ \\
\midrule
\multicolumn{2}{l}{\textit{Cross-template}: $i_s \sim t_1, i_t \sim t_2$} \\
\quad Answer-only: $(i_s, a_m(i_s), i_t)$ & $a_m(i_t)$ \\
\quad With-steps: $(i_s, a_m(i_s), S_m(i_s), i_t)$ & $a_m(i_t)$ \\
\bottomrule
\end{tabular}
\caption{MRA experimental conditions.}
\label{tab:mra_conditions}
\end{table}

\paragraph{Forward MRA Task.} We also evaluate whether models can apply a misconception given its description. Let $D(m)$ denote a natural language description of malrule $m$. Given $D(m)$ and a problem $i$, the model must predict $a_m(i)$. Table~\ref{tab:prompt_comparison} illustrates both MRA and Forward MRA prompt formats.

\paragraph{Correct Reasoning Accuracy (CRA).} To contextualize the difficulty of MRA and Forward MRA, we also evaluate whether models can solve problems correctly. Given a problem $i$, the model must predict $a_c(i)$, the correct answer.

\begin{table*}[t]
\centering
\small
\begin{tabular}{l|c|cc|cccc|cccc}
\toprule
& & \multicolumn{2}{c|}{FMRA} & \multicolumn{4}{c|}{MRA (no steps)} & \multicolumn{4}{c}{MRA (w/ steps)} \\
Model & CRA & Acc & $\Delta$ & Same & $\Delta$ & Cross & $\Delta$ & Same & $\Delta$ & Cross & $\Delta$ \\
\midrule
gpt-oss-20b & 65.2 & 46.5 & -18.7 & 72.7 & +7.5 & 53.6 & -11.6 & 78.5 & +13.2 & 57.7 & -7.6 \\
Qwen3-4B & 67.3 & 17.4 & -49.9 & 61.1 & -6.2 & 39.1 & -28.2 & 55.7 & -11.6 & 34.6 & -32.7 \\
Phi-4 & 64.9 & 37.4 & -27.5 & 50.0 & -14.9 & 36.7 & -28.2 & 64.9 & -0.0 & 47.2 & -17.7 \\
Phi-4-mini & 55.2 & 9.3 & -45.9 & 26.7 & -28.5 & 18.2 & -37.0 & 42.5 & -12.7 & 26.3 & -28.9 \\
Llama-3.1-8B & 63.8 & 6.7 & -57.1 & 25.3 & -38.5 & 17.7 & -46.1 & 43.6 & -20.1 & 29.4 & -34.4 \\
\midrule
\textit{Small avg} & 63.3 & 23.5 & -39.8 & 47.2 & -16.1 & 33.1 & -30.2 & 57.1 & -6.2 & 39.0 & -24.3 \\
\midrule
gpt-oss-120b & 65.0 & 48.8 & -16.1 & 77.1 & +12.1 & 56.9 & -8.1 & 81.5 & +16.5 & 60.9 & -4.1 \\
Qwen3-80B-Think & 70.1 & 53.9 & -16.2 & 74.2 & +4.1 & 56.4 & -13.7 & 77.2 & +7.1 & 59.8 & -10.3 \\
Qwen3-80B-Inst & 69.8 & 30.6 & -39.2 & 69.9 & +0.1 & 51.3 & -18.5 & 73.2 & +3.4 & 54.4 & -15.4 \\
Llama-3.3-70B & 70.4 & 39.6 & -30.7 & 47.7 & -22.7 & 34.6 & -35.8 & 64.4 & -5.9 & 48.5 & -21.9 \\
\midrule
\textit{Large avg} & 68.8 & 43.2 & -25.6 & 67.2 & -1.6 & 49.8 & -19.0 & 74.1 & +5.3 & 55.9 & -12.9 \\
\midrule
\textbf{Overall} & \textbf{65.7} & \textbf{32.3} & \textbf{-33.5} & \textbf{56.1} & \textbf{-9.7} & \textbf{40.5} & \textbf{-25.3} & \textbf{64.6} & \textbf{-1.1} & \textbf{46.5} & \textbf{-19.2} \\
\bottomrule
\end{tabular}
\caption{Performance across experimental conditions. CRA = problem solving. FMRA = applying described misconception. MRA = predicting student answer from example. Same/Cross = template generalization. $\Delta$ = gap from CRA (negative = below CRA). Models grouped by size, sorted by Cross MRA. Across models, CRA exceeds MRA by large margins, and the gap widens under cross-template evaluation. Providing step-by-step reasoning traces typically improves MRA.}
\label{tab:consolidated}
\end{table*}

\subsection{Sampled Dataset Statistics}

We sample ${\sim}10$ instances per template from all 101 malrules, yielding 4,991 problem instances across 498 (malrule, template) groups. For \textit{same-template} pairs, we sample $(i_s, i_t)$ from the same group, selecting up to 10 pairs per group. For \textit{cross-template} pairs, we sample $i_s \sim t_1$ and $i_t \sim t_2$ where $t_1 \neq t_2$, selecting 100 pairs per malrule (77 malrules have $|\mathcal{T}_m| \geq 2$). This yields 12,706 pairs: 5,006 same-template and 7,700 cross-template.

\textbf{Inference calls.} For MRA, each of the 12,706 pairs is evaluated under two prompt conditions (answer-only and with-steps), yielding ${\sim}25$K calls per model. For CRA, each of the 4,991 problem instances is solved once, yielding ${\sim}5$K calls. For Forward MRA, each instance is evaluated with its malrule description, yielding another ${\sim}5$K calls. In total, each model requires ${\sim}35$K inference calls. With 9 models evaluated, the benchmark comprises ${\sim}320$K total calls.

\subsection{Models and Evaluation}

We evaluate nine language models spanning two size categories. Large models (70--120B parameters) are gpt-oss-120b \cite{agarwal2025gpt}, Qwen3-80B-Think \cite{yang2025qwen3}, Qwen3-80B-Instruct, and Llama-3.3-70B \cite{touvron2023llama}. Small models (4--20B parameters) are gpt-oss-20b, Qwen3-4B, Phi-4 \cite{abdin2024phi}, Phi-4-mini, and Llama-3.1-8B. Following model card recommendations, we use the following sampling parameters: for Qwen3 models in thinking mode, temperature 0.6, top-$p$ 0.95, and top-$k$ 20; for Qwen3 in non-thinking mode, temperature 0.7 and top-$p$ 0.8; and for gpt-oss models, temperature 1.0 and top-$p$ 1.0. Our primary metric is \textbf{Malrule Reasoning Accuracy (MRA)} to measure performance of malrule reasoning task: whether the model predicts the specific wrong answer produced by the malrule. We use normalized matching for algebraic expressions and numerical matching with tolerance for decimal answers.

\section{Results and Discussion}

\begin{figure*}[t]
\centering
\begin{minipage}[b]{0.72\textwidth}
    \centering
    \includegraphics[width=\textwidth]{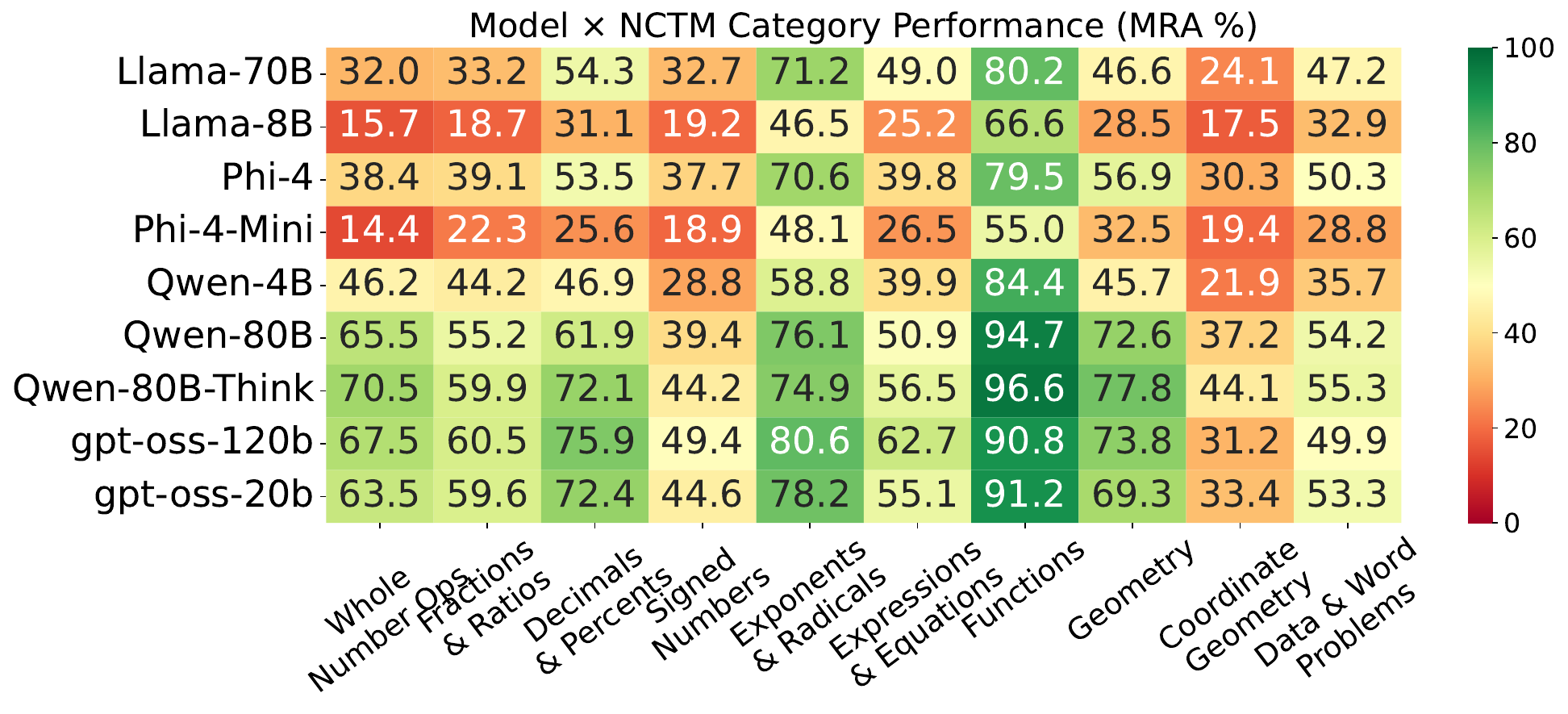}
    \captionof{figure}{Performance by mathematical category. Functions is easiest (82\%), Coordinate Geometry hardest (29\%).}
    \label{fig:category}
\end{minipage}
\hfill
\hfill
\begin{minipage}[b]{0.55\textwidth}
    \centering
    {\small
\setlength{\tabcolsep}{2pt}
\begin{tabular}{llr}
\toprule
Category & Strand & MRA \\
\midrule
Whole Number Ops & Number \& Operations & 46.0 \\
Fractions \& Ratios & Number \& Operations & 43.6 \\
Decimals \& Percents & Number \& Operations & 54.9 \\
Signed Numbers & Number \& Operations & 35.0 \\
Exponents \& Radicals & Algebra & 67.2 \\
Expressions \& Equations & Algebra & 45.1 \\
Functions & Algebra & 82.1 \\
Geometry & Geometry \& Measurement & 56.0 \\
Coordinate Geometry & Geometry \& Measurement & 28.8 \\
Data \& Word Problems & Data \& Modeling & 45.3 \\
\bottomrule
\end{tabular}}

    \captionof{table}{Average MRA by NCTM category across all models.}
    \label{tab:category_breakdown}
\end{minipage}
\end{figure*}

Table~\ref{tab:consolidated} summarizes performance on three tasks: \textbf{CRA} (solve correctly), \textbf{MRA} (infer a student's malrule from one example and predict the next answer), and \textbf{Forward MRA (FMRA)} (apply a described misconception). We additionally separate \textbf{Same} versus \textbf{Cross} template evaluation to measure generalization, and we compare \textbf{answer-only} versus \textbf{with-steps} evidence to quantify the value of reasoning traces. Figure~\ref{fig:category} and Table~\ref{tab:category_breakdown} break results down by mathematical domain.

\subsection{Capability gaps in student modeling}

\paragraph{MRA remains below CRA.}
Models solve problems far better than they predict misconception-driven answers. Overall CRA is 65.7\%, while cross-template MRA is only 40.5\% (answer-only) and 46.5\% (with steps), leaving gaps of 25.3 and 19.2 points  (Table~\ref{tab:consolidated}). This gap captures a core barrier for tutoring: competence at producing correct mathematics does not imply competence at modeling how a student will systematically be wrong.

A key reason is that MRA is a different computation than CRA. CRA is a single forward pass toward correctness, whereas MRA requires both inferring a latent procedure (the student's malrule) from limited evidence and then executing that same flawed procedure on a new instance. This counterfactual execution conflicts with strong training priors for correct, helpful answers and is consistent with a supervision imbalance: models see extensive curated data for correct reasoning during pretraining and instruction tuning, but far less curated data where an incorrect procedure is executed consistently and treated as the intended output.

Finally, note that same-template MRA can exceed CRA for some models, indicating that reproducing a local error pattern within a template family can be easier than solving the underlying mathematics. This is another reason cross-template evaluation is essential.

\paragraph{FMRA remains below CRA.}
Forward MRA is substantially below CRA: overall FMRA is 32.3\% versus CRA at 65.7\% (Table~\ref{tab:consolidated}). In FMRA, the misconception is explicitly stated, so the failure is not lack of access to the rule. Rather, it reflects difficulty converting a natural-language description into a faithful, repeatable algorithmic transformation. Forward MRA requires translating a high-level misconception statement into a concrete sequence of algebraic operations. Phrases such as ``distribute,'' ``cancel,'' or ``add inside'' require the model to decide exactly where and how the flawed operation applies. In addition, instruction tuning encourages models to correct misconceptions rather than simulate them, creating a tension between being correct and behaving like a mistaken student.

\paragraph{Examples outperform descriptions.}
Despite giving the misconception explicitly, FMRA is generally worse than example-based MRA: overall FMRA is 32.3\%, while cross-template MRA reaches 40.5 to 46.5\% depending on whether steps are provided (Table~\ref{tab:consolidated}). This pattern suggests that examples provide operational semantics that align with LLM strengths. A single worked mistake implicitly defines the mapping from problem structure to erroneous output, enabling in-context induction of the malrule.
This suggests that executed examples with solution traces may be a more effective supervision format than descriptions alone for improving misconception application.

\subsection{Value of step evidence and generalization}

\paragraph{Reasoning traces help most models.}
\textsc{MalruleLib} provides full step-by-step solution traces for both correct reasoning and malrule-based student reasoning, and these traces translate into measurable gains in student modeling. On average, cross-template MRA rises from 40.5\% to 46.5\% when steps are included, and many individual models see improvements (Table~\ref{tab:consolidated}).
This result validates our design decision to generate dual-path reasoning traces. It also points to a concrete training direction: supervised fine-tuning on malrule-consistent step traces can teach models to execute them faithfully under cross-template shifts.

\paragraph{Cross-template generalization is the bottleneck.}
Another core contribution of \textsc{MalruleLib} is that each malrule is instantiated across many diverse templates, enabling a direct test of whether models represent misconceptions as abstract procedures rather than template cues. Under this cross-template setting, performance drops sharply. Overall, same-template MRA (answer-only) is 56.1\% while cross-template is 40.5\%. With steps, the same-template score is 64.6\% while cross-template is 46.5\% (Table~\ref{tab:consolidated}). Every model degrades when the surface form changes, indicating that many same-template successes can be achieved by template-level pattern matching rather than an abstract representation of the malrule.
For example, cross-template items include shifts where the misconception must be applied inside a different wrapper, such as moving from a direct radical simplification to a distance word problem that implicitly requires $\sqrt{a^2+b^2}$. These are precisely the settings where a tutor must recognize and anticipate the same flawed procedure across contexts.

\paragraph{Large domain spread.}
Performance varies sharply by mathematical category. Functions is easiest (82.1\%) while Coordinate Geometry is hardest (28.8\%), with Signed Numbers also low (35.0\%), yielding a 53-point spread (Table~\ref{tab:category_breakdown}). Figure~\ref{fig:category} shows this pattern is consistent across model families.
The spread in Table~\ref{tab:category_breakdown} suggests that student modeling is not monolithic and should be validated by domain before deployment. In practice, tutoring systems may require fine-tuning for certain categories such as geometry and signed arithmetic.

\section{Conclusion}

We introduced \textsc{MalruleLib}, a learning-science-grounded framework that encodes 101 documented mathematical malrules over 498 problem templates and generates dual-path step-by-step traces for both correct reasoning and malrule-consistent student reasoning. Using \textsc{MalruleLib}, we built the first large-scale benchmark for misconception-based student modeling with controlled same-template versus cross-template evaluation.
Across nine language models (4B--120B), we find a persistent gap between problem solving and predicting misconception-driven answers, with substantial cross-template degradations. Providing student work yields consistent gains, motivating evaluation protocols that test cross-context generalization and motivating educational systems to capture intermediate reasoning, not only final answers.
Because malrules are executable and templates are parameterizable, \textsc{MalruleLib} can generate large-scale training data with malrule-consistent intermediate steps for fine-tuning and instruction tuning. We release \textsc{MalruleLib} as infrastructure for student modeling, a core capability for personalized learning, tutoring, and feedback at scale.

\bibliographystyle{acl_natbib}
\bibliography{references}

\clearpage
\appendix
\section*{Appendix}

\section{Additional Malrule Examples}
Table~\ref{tab:malrule_ex_examples} expands on Table~\ref{tab:prompt_comparison} with additional malrule examples across 10 categories, each showing two template variations demonstrating cross-template generalization.

\begin{table*}[t]
\centering
\small
\begin{tabular}{p{1.8cm}p{2.5cm}p{5.5cm}p{4.5cm}}
\toprule
\textbf{Category} & \textbf{Malrule} & \textbf{Problem} & \textbf{Student's Work} \\
\midrule
\multirow{2}{*}{Statistics} & \multirow{2}{*}{Mode must exist}
& \textbf{T1:} Identify the mode for the following data set: [15, 95, 36, 15, 4, 82]
& Frequencies: 4:1, 15:2, 36:1, 82:1, 95:1 $\rightarrow$ 95 \\
& & \textbf{T2:} Dataset A: [2, 8, 41, 41], Dataset B: [68, 65, 52, 66, 91, 58]. Which dataset has a mode?
& Dataset A: Mode = 41(correct), Dataset B: Mode = 91(picked biggest) \\
\midrule
\multirow{2}{*}{Algebra} & \multirow{2}{*}{\shortstack[l]{$x + a = b \rightarrow$ \\ $x = b + a$}}
& \textbf{T1:} Solve for x: $4x + 9 - 7 = 34$
& $4x + 9 - 7 = 34 \rightarrow 4x = 34 + 9 - 7$; $\rightarrow 4x = 36 \rightarrow x = 9$ \\
& & \textbf{T2:} A phone plan costs \$29 monthly plus \$5 per GB of data. If the bill is \$114, how many GB (x) were used?
& $5x + 29 = 114 \rightarrow 5x = 114 + 29$ $ \rightarrow 5x = 143 \rightarrow x = 28$ \\
\midrule
\multirow{2}{*}{\shortstack[l]{Scientific \\Notation}} & \multirow{2}{*}{\shortstack[l]{Count all zeros \\ for exponents}}
& \textbf{T1:} Write 0.010500 in scientific notation.
& Zeros count: 5, Coefficient: 1.05 $\rightarrow$ Exponent: -5, $1.05 \times 10^{-5}$ \\
& & \textbf{T2:} Is 0.00002050 equal to $2.05 \times 10^{-7}$?
& Zeros count: 7, Coefficient: 2.05, Exponent: -7, Answer: Yes \\
\midrule
\multirow{2}{*}{Absolute Value} & \multirow{2}{*}{|a+b| = |a|+|b|}
& \textbf{T1:} Evaluate |x - 2| when x = 10
& |x - 2| = |x| - 2 = 10 - 2 = 8 \\
& & \textbf{T2:} A drone is at position x meters, and moves 3 meters to the right. The distance from the origin is |x + 3|. If x = -3, what is the distance from the origin?
& |x + 3| = |x| + 3 = 3 + 3 = 6 meters \\
\midrule
\multirow{2}{*}{Decimals} & \multirow{2}{*}{\shortstack[l]{More digits \\ $\rightarrow$ larger value}}
& \textbf{T1:} Which is longer: 0.5 kilometers or 0.479 kilometers?
& 0.5: 1 places, 0.479: 3 places, $3 > 1$, 0.479 \\
& & \textbf{T2:} Maria has \$0.61 and Tom has \$0.214. Who has more money?
& 0.61: 2 places, 0.214: 3 places, $3 > 2$, Tom \\
\midrule
\multirow{2}{*}{Exponents} & \multirow{2}{*}{$(a+b)^n = a^n + b^n$}
& \textbf{T1:} Evaluate: $(1 + 3 + 2)^2$
& $(1 + 3 + 2)^2 = 1^2 + 3^2 + 2^2 = 1 + 9 + 4 = 14$ \\
& & \textbf{T2:} A server processes 2 GB in phase 1 and 4 GB in phase 2 (total: 6 GB). This data volume is replicated across 2 redundant systems with exponential scaling. Calculate $(2+4)^2$.
& $(2 + 4)^2 = 2^2 + 4^2 = 4 + 16 = 20$\\
\midrule
\multirow{2}{*}{Factoring} & \multirow{2}{*}{$a^2 + b^2 = (a+b)^2$}
& \textbf{T1:} Factor: $x^2 + 36$
& $(x + 6)^2 = x^2 + 36$, $(x + 6)^2$ \\
& & \textbf{T2:} Factor: $3^2x^2 + 5^2y^2$
& $(3x + 5y)^2 = 3^2x^2 + 5^2y^2$, $(3x + 5y)^2$ \\
\midrule
\multirow{2}{*}{Fractions} & \multirow{2}{*}{\shortstack[l]{Add numerators \\ together and \\denominators together}}
& \textbf{T1:} What is $\frac{5}{3} + \frac{7}{4}$?
& $\frac{5}{3} + \frac{7}{4}$, 5 + 7 = 12, 3 + 4 = 7 $\rightarrow$ $\frac{12}{7}$ \\
& & \textbf{T2:} Sarah ate $\frac{1}{4}$ of a pizza and John ate $\frac{1}{3}$ of the same pizza. What fraction of the pizza did they eat together??
& $\frac{1}{4} + \frac{1}{3}$, 1 + 1 = 2, 4 + 3 = 7 $\rightarrow$ $\frac{2}{7}$ \\
\midrule
\multirow{2}{*}{Geometry} & \multirow{2}{*}{\shortstack[l]{Surface area = \\$l \times w \times h$}}
& \textbf{T1:} Find the surface area of a rectangular prism with length 4.1 cm, width 5.4 cm, and height 3.0 cm.
& A = $l \times w \times h$ = 4.1 × 5.4 × 3.0 = 66.42 \\
& & \textbf{T2:} A storage container measures 4 feet long, 8 feet wide, and 8 feet tall. What is the total surface area that needs to be painted?
& A = $l \times w \times h$ = 4 × 8 × 8 = 256 \\
\midrule
\multirow{2}{*}{\shortstack[l]{Linear \\Equations}} & \multirow{2}{*}{\shortstack[l]{Slope = $\frac{\Delta x}{\Delta y}$}}
& \textbf{T1:} Find the slope of the line passing through points (-9, 2) and (1, 8).
& $\Delta x = 10$, $\Delta y = 6$, Slope = $\frac{10}{6} = \frac{5}{3}$ \\
& & \textbf{T2:} After 15 hours, a vehicle has gone 18 miles. After 19 hours, it has gone 5 miles. Calculate the speed.
& Point 1 = (15, 18), Point 2 = (19, 5), $\Delta x = 4$, $\Delta y = -13$, Slope = $\frac{4}{-13} = -\frac{4}{13}$ \\\midrule
\bottomrule
\end{tabular}
\caption{More Examples of malrules from \textsc{MalruleLib}, showing two templates per misconception.}
\label{tab:malrule_ex_examples}
\end{table*}

\section{Malrules and Their Sources}
Tables~\ref{tab:malrulesoc}--\ref{tab:malrulesoc4} present the source and description for each malrule in this study.

\begin{table*}[t]
\small
\centering
\begin{tabular}{p{0.45\textwidth} p{0.55\textwidth}}
\toprule
\textbf{Malrule (Source)} & \textbf{Description} \\
\midrule
absolute\_value\_distributes \cite{M1M2} & treating $|a+b|$ as $|a|+|b|$ \\
absolute\_value\_makes\_positive \cite{M1M2} & Students hold the mental picture that 'absolute values make negative signs positive' and apply this symbol-based process rather than understanding that absolute value represents distance from zero, treats $|x| = -a$ as $x = a$ \\
inequality\_direction\_confusion \cite{M3} & Students confuse the solution patterns for absolute value inequalities, applying the 'less than' pattern (three-part-inequality) to 'greater than' problems and vice versa. \\
cancel\_across\_equals \cite{M4M6} & Students incorrectly 'cancel' matching variable terms from both sides of an equation as if they were canceling factors in a fraction, rather than properly subtracting the terms from both sides. \\
change\_side\_change\_sign \cite{M5} & Students incorrectly believe that when moving a term from one side of an equation to the other, no sign need to change, ie., $x + a = b \rightarrow x = b + a$ \\
distribute\_over\_non\_distributive \cite{M4M6} & Students incorrectly extend the distributive property to operations that do not distribute, such as exponentiation and square roots. \\
divide\_one\_term\_only \cite{M7} & Students incorrectly apply division to only one term (typically the variable term) instead of distributing the operation to all terms on that side of the equation. \\
forget\_negative\_division \cite{M8} & Students neglect the crucial rule that multiplying or dividing an inequality by a negative coefficient reverses the direction of the inequality sign. \\
variable\_letter\_has\_value \cite{M9} & Students believe that algebraic letters have inherent fixed values, often based on the letter's position in the alphabet (e.g., $x=10$, $a=1$) or other associations. \\
ignore\_decimal\_point \cite{M10M14} & Students ignore the decimal point and treat decimal numbers as whole numbers, applying whole number arithmetic procedures without regard to place value. \\
longer\_is\_larger \cite{M11} & When comparing decimal numbers, students incorrectly believe that more decimal digits means a larger number. \\
right\_align\_decimals \cite{M12} & Students align decimal numbers by the rightmost digit instead of by the decimal point when performing addition or subtraction. \\
shorter\_is\_larger \cite{M13} & Students incorrectly believe that decimals with fewer decimal places are larger. \\
whole\_number\_thinking \cite{M10M14} & Students incorrectly treat decimals as if they demonstrate the properties of whole numbers. For example, $3.7 + 2.5$ treated as $37 + 25 = 62$, $0.45 > 0.8$ because $45 > 8$ \\
add\_exponents\_for\_power\_of\_power \cite{M15} & Student adds exponents instead of multiplying them when computing power of a power. \\
distribute\_exponent\_over\_addition \cite{M16} & Student incorrectly distributes exponents over addition: $(a+b)^n = a^n + b^n$. \\
forget\_exponent\_on\_coefficient \cite{M17} & Student forgets to raise coefficient to the outer power: $(cx^m)^n = c \cdot x^{mn}$. \\
multiply\_base\_by\_exponent \cite{M18} & Student treats exponentiation as multiplication: $a^b = a \times b$. \\
multiply\_exponents\_when\_multiplying\_powers \cite{M19} & Student multiplies exponents when multiplying powers with the same base: $x^m \times x^n = x^{m \times n}$. \\
negative\_exponent\_makes\_negative \cite{M20} & Student thinks negative exponent makes the result negative: $x^{-n} = -x^n$. \\
zero\_exponent\_equals\_zero \cite{M21} & Student thinks $a^0 = 0$ instead of $a^0 = 1$. \\
incomplete\_factoring \cite{M22} & Stop after first factoring step when the result is not fully factored \\
\bottomrule
\end{tabular}
\caption{Source and description for each malrule (Part 1 of 4).}
\label{tab:malrulesoc}
\end{table*}

\begin{table*}[t]
\small
\centering
\begin{tabular}{p{0.45\textwidth} p{0.55\textwidth}}
\toprule
\textbf{Malrule (Source)} & \textbf{Description} \\
\midrule
sign\_errors\_in\_factoring \cite{M23} & Students make sign errors when factoring quadratics, especially in decomposition method. Such as $a^2 - b^2 = (a-b)(a+b)$, perfect square, and factoring constants sign. \\
sum\_of\_squares\_factors \cite{M24} & Student thinks $a^{2} + b^{2} = (a+b)^{2}$ \\
add\_numerators\_add\_denominators \cite{M25} & When adding fractions with different denominators, students incorrectly add numerators together and denominators together, treating the operation as component-wise addition. \\
common\_denominator\_numerator \cite{M26} & This is a variant of the add-numerators-add-denominators error that shows partial understanding - students know they need a common denominator but incorrectly believe adding denominators produces one. \\
denominator\_comparison\_error \cite{M27} & Students incorrectly compare fractions by focusing on the denominator value, believing that a larger denominator means a larger fraction. \\
ignore\_denominators \cite{M28} & Students operate only on numerators, completely ignoring denominators, treating fraction operations as whole-number arithmetic on the 'top numbers' only. \\
keep\_common\_denominator\_for\_multiplication \cite{M29} & Students incorrectly keep the common denominator when multiplying fractions with like denominators: $(a/b) \times (c/b) = (a \times c)/b$ instead of $(a \times c)/(b \times b)$. \\
multiply\_across\_for\_division \cite{M30} & Student treats fraction division like multiplication, forgetting to invert (flip) the second fraction before multiplying. \\
natural\_number\_bias\_numerator\_only \cite{M31} & Students incorrectly compare fractions by comparing their numerators only, ignoring the denominators entirely. They believe that a larger numerator means a larger fraction. \\
subtract\_across \cite{M32} & When subtracting fractions with different denominators, students incorrectly subtract the numerators AND subtract the denominators: $\frac{a}{b} - \frac{c}{d} = (a-c)/(b-d)$. \\
function\_distributive\_property \cite{M33} & Student incorrectly applies the additive property of linear functions to nonlinear functions: $f(x+a) = f(x) + f(a)$. \\
function\_notation\_is\_multiplication \cite{M34} & Student interprets $f(x)$ as meaning $f \times x$ (multiplication) rather than function notation. \\
same\_input\_different\_outputs\_ok \cite{M35} & Students fail to understand the univalence requirement: that functions must map each input to exactly ONE output. \\
scalar\_multiplication\_inside\_or\_outside\_same \cite{M36} & Student incorrectly applies the multiplicative (homogeneous) property of linear functions to nonlinear functions: $f(cx) = c \cdot f(x)$. \\
count\_net\_perimeter\_as\_surface\_area \cite{M37} & Measure net perimeter instead of calculating surface area \\
same\_area\_same\_perimeter \cite{M38M39} & Students think same area means same perimeter. \\
same\_perimeter\_same\_area \cite{M38M39} & Students think same perimeter means same area. \\
volume\_formula\_for\_surface\_area \cite{M40} & Students use volume formula $V=lwh$ when asked for surface area. \\
ignore\_coordinate\_signs \cite{M41} & Student ignores or misuses the signs of coordinates, treating negative values as positive (absolute values) when plotting points or identifying quadrants. \\
confuse\_slope\_and\_intercept\_roles \cite{M42M43} & Student swaps the roles of slope ($m$) and y-intercept ($b$) in $y = mx + b$, either writing equations with parameters reversed or identifying the constant term as slope and the coefficient as y-intercept. \\
slope\_direction\_confusion \cite{M42M43} & Student confuses positive and negative slope directions, incorrectly stating that negative slopes represent increasing functions or positive slopes represent decreasing functions. \\
slope\_is\_delta\_x\_over\_delta\_y \cite{M44} & Student inverts the slope formula, calculating $\frac{\Delta x}{\Delta y}$ instead of $\frac{\Delta y}{\Delta x}$. \\
\bottomrule
\end{tabular}
\caption{Source and description for each malrule (Part 2 of 4).}
\label{tab:malrulesoc2}
\end{table*}

\begin{table*}[t]
\small
\centering
\begin{tabular}{p{0.45\textwidth} p{0.55\textwidth}}
\toprule
\textbf{Malrule (Source)} & \textbf{Description} \\
\midrule
inverted\_conversion\_factor \cite{M45M46} & Students set up the conversion factor incorrectly by placing units in the wrong position (numerator vs. denominator), resulting in multiplication when division is required or vice versa. \\
wrong\_conversion\_factor \cite{M45M46} & Students use incorrect conversion factors when converting between units, often substituting convenient round numbers (10, 100) or powers of ten for the actual conversion factors. \\
alignment\_error\_in\_multi\_digit \cite{M47M50} & When multiplying by the tens digit, students do not shift the partial product one place to the left, treating the tens digit as if it were in the ones place. \\
divide\_larger\_by\_smaller\_always \cite{M48M49M51} & A persistent misconception where students believe division must always involve a larger dividend divided by a smaller divisor. When presented with problems where the dividend is smaller than the divisor (e.g., 4 ÷ 6), students either reverse the operands or claim the problem is impossible. \\
division\_makes\_smaller \cite{M48M49M51} & A widespread misconception where students believe division always produces a result smaller than the dividend. \\
forget\_to\_add\_carried\_number \cite{M47M50} & A systematic procedural bug where students correctly multiply each digit but forget to add the carried (regrouped) value to the next place value. \\
multiplication\_makes\_bigger \cite{M48M49M51} & A widespread misconception where students believe multiplication always produces a result larger than both factors. \\
larger\_absolute\_value\_always\_wins \cite{M52M53M56} & Students incorrectly determine the sign of the result by always using the sign of the number with the larger absolute value, regardless of the operation. \\
multiplication\_rule\_for\_addition \cite{M52M53M56} & Students incorrectly apply multiplication sign rules (like 'two negatives make a positive') to addition and subtraction operations. \\
negative\_swaps\_operation \cite{M54} & Students confuse the negative sign's role as a number property with the operation being performed, leading to incorrect operation swapping. \\
negative\_times\_negative\_negative \cite{M55} & Students incorrectly believe that multiplying two negative numbers gives a negative result, reversing the correct sign rule. \\
two\_negatives\_always\_positive \cite{M52M53M56} & Students incorrectly apply the multiplication rule 'two negatives make a positive' to subtraction and addition operations. \\
addition\_before\_subtraction-always \cite{M57M59} & Students incorrectly believe that addition must ALWAYS be performed before subtraction. \\
ignore\_parentheses \cite{M58} & Students ignore parentheses and evaluate left-to-right or apply PEMDAS without respecting grouping symbols. \\
multiplication\_before\_division\_always \cite{M57M59} & Students incorrectly believe that multiplication must ALWAYS be performed before division. \\
pemdas\_strictly\_sequential \cite{M60} & Students treat PEMDAS as a strict six-step sequence rather than understanding it as four priority levels. \\
strict\_left\_to\_right \cite{M61} & Students evaluate arithmetic expressions strictly from left to right, ignoring operator precedence rules (PEMDAS/BODMAS). \\
add\_percentages\_directly \cite{M62} & Students incorrectly apply additive reasoning to percentage problems that require multiplicative thinking. They add or subtract percentages directly without recognizing that each percentage applies to a different base value. \\
percent\_equals\_decimal \cite{M63M64M65} & Students treat percent, decimal, and whole number notations as interchangeable, failing to recognize that percent means 'per hundred' and requires division by 100 to convert to decimal form. \\
percentage\_as\_index \cite{M63M64M65} & Students treat the percentage number as an absolute value or index, ignoring the base/whole that the percentage applies to. \\
reverse\_percentage\_error \cite{M63M64M65} & Students incorrectly believe that percentage relationships are symmetric or reversible. They assume that if a value increased by X\%, it can return to the original by decreasing X\%, or that if A is X\% of B, then B is X\% of A. \\
add\_under\_common\_root \cite{M66M67} & Students combine radicals under common root, i.e., students believes that $\sqrt{a} - \sqrt{b} = \sqrt{(a-b)}$. \\
\bottomrule
\end{tabular}
\caption{Source and description for each malrule (Part 3 of 4).}
\label{tab:malrulesoc3}
\end{table*}

\begin{table*}[t]
\small
\centering
\begin{tabular}{p{0.45\textwidth} p{0.55\textwidth}}
\toprule
\textbf{Malrule (Source)} & \textbf{Description} \\
\midrule
distribute\_square\_root\_over\_addition \cite{M66M67} & Students believe that square root distributes over addition, $\sqrt{}$ applies to each term separately, i.e. $\sqrt{a^2 + b^2} = \sqrt{a^2} + \sqrt{b^2}$ \\
negative\_outside\_same\_as\_inside \cite{M68} & Students confuse $-\sqrt{n}$ with $\sqrt{-n}$, failing to recognize that the domain of the radical function is restricted to nonnegative real numbers. \\
square\_root\_equals\_plus\_minus \cite{M69} & Students incorrectly believe that the radical symbol $\sqrt{n}$ yields both positive and negative values ($\pm\sqrt{n}$), confusing the principal square root with the solutions to equations of the form $x^2 = n$. \\
additive\_instead\_of\_multiplicative \cite{M70} & Student applies additive reasoning instead of multiplicative reasoning when working with proportions. The student computes the difference between values and adds this constant to find missing values, rather than using the multiplicative scale factor. \\
each\_fraction\_digit\_is\_ratio \cite{M71} & Student treats the numerator and denominator as independent whole numbers rather than as components of a single rational number. \\
ratio\_as\_division\_only \cite{M72M73} & Student interprets ratio solely as division (a quotient), failing to understand that a ratio represents a multiplicative comparison between two quantities. \\
swap\_ratios\_or\_units \cite{M72M73} & Student incorrectly sets up proportion equations by placing values in wrong positions, inverting the relationship between quantities. \\
decimal\_places\_same\_as\_sig\_figs \cite{M74} & Students confuse decimal places with significant figures. When asked to round to N significant figures, they instead round to N decimal places. \\
add\_coefficients\_when\_multiplying \cite{M75M77} & When multiplying numbers in scientific notation, students correctly add the exponents but incorrectly add the coefficients instead of multiplying them. \\
count\_all\_zeros\_for\_exponent \cite{M76} & Students count ALL zeros in a number when determining the exponent for scientific notation, without considering their placement. \\
wrong\_exponent\_sign \cite{M75M77} & Students use the wrong sign for the exponent when converting numbers to scientific notation. \\
ignore\_outliers\_effect \cite{M78M79} & Student tend to ignore outliers and assume mean is always representative \\
mean\_without\_understanding \cite{M78M79} & Student always use mean as the primary measure regardless the case. \\
mode\_must\_exist \cite{M80} & Students force a mode to exist by picking largest/middle/smallest value in the data. \\
always\_borrow\_left \cite{M81M83M85M87M89} & Students always borrow from the left column in subtraction, even when the top digit is greater than or equal to the bottom digit and borrowing is not needed. \\
borrow\_from\_bottom \cite{M82} & When borrowing is needed, students incorrectly decrement the subtrahend (bottom number) instead of the minuend (top number). \\
borrow\_no\_decrement \cite{M81M83M85M87M89} & When borrowing is required, students correctly add 10 to the current column's digit but forget to decrement the digit in the column they borrowed from. \\
decompose\_by\_place\_value\_label \cite{M84} & When interpreting place value decompositions, the student treats the numeric labels as face values to be concatenated rather than as multiplicative values to be added. \\
diff\_0\_n\_equals\_n \cite{M81M83M85M87M89} & When the minuend digit is 0 and the subtrahend digit is N (where N > 0), the student writes N as the result instead of borrowing. \\
no\_column\_limit \cite{M86} & When adding multi-digit numbers, students write the entire column sum directly in that column without regrouping, failing to understand the base-10 constraint that each place value position can only hold a single digit (0-9). \\
smaller\_from\_larger \cite{M81M83M85M87M89} & A systematic procedural bug where the student always subtracts the smaller digit from the larger digit in each column, regardless of position (minuend or subtrahend). \\
stops\_borrow\_at\_zero \cite{M88} & A systematic procedural bug where the student stops the borrowing (regrouping) process entirely when encountering a zero in the column to the left, rather than cascading the borrow further left to find a non-zero digit. \\
\bottomrule
\end{tabular}
\caption{Source and description for each malrule (Part 4 of 4).}
\label{tab:malrulesoc4}
\end{table*}

\section{Full Results}

\subsection{Per-Malrule Breakdown}
Tables~\ref{tab:per_malrule}--\ref{tab:per_malrule3} present performance breakdown for all 101 malrules, sorted by accuracy.

\begin{table*}[t]
\centering
\small
\begin{tabular}{llll}
\toprule
Malrule & Total & MRA & Accuracy \\
\midrule
subtraction.borrow\_from\_bottom & 2700 & 132 & 4.89 \\
algebra.variable\_letter\_has\_value & 2700 & 169 & 6.26 \\
fractions.multiply\_rule\_for\_addition & 180 & 22 & 12.22 \\
ratios\_proportions.each\_fraction\_digit\_is\_ratio & 2700 & 331 & 12.26 \\
statistics.mean\_without\_understanding & 3600 & 546 & 15.17 \\
subtraction.carry\_ones\_digit\_instead\_of\_tens & 2700 & 544 & 20.15 \\
absolute\_value.absolute\_value\_makes\_positive & 2700 & 551 & 20.41 \\
negative\_numbers.negative\_times\_negative\_negative & 3240 & 750 & 23.15 \\
scientific\_notation.wrong\_exponent\_sign & 2700 & 632 & 23.41 \\
measurement.wrong\_conversion\_factor & 180 & 43 & 23.89 \\
radicals.add\_under\_common\_root & 3420 & 862 & 25.20 \\
scientific\_notation.count\_all\_zeros\_for\_exponent & 2700 & 696 & 25.78 \\
fractions.natural\_number\_bias\_numerator\_only & 3060 & 836 & 27.32 \\
factoring.sign\_errors\_in\_factoring & 180 & 50 & 27.78 \\
negative\_numbers.two\_negatives\_always\_positive & 2700 & 750 & 27.78 \\
graphing.reverse\_coordinate\_order & 180 & 51 & 28.33 \\
graphing.ignore\_coordinate\_signs & 2700 & 778 & 28.81 \\
algebra.divide\_one\_term\_only & 180 & 52 & 28.89 \\
subtraction.smaller\_from\_larger & 4140 & 1234 & 29.81 \\
subtraction.no\_column\_limit & 2700 & 809 & 29.96 \\
algebra.change\_side\_change\_sign & 180 & 55 & 30.56 \\
order\_of\_operations.addition\_before\_subtraction\_always & 2700 & 829 & 30.70 \\
order\_of\_operations.ignore\_parentheses & 2700 & 836 & 30.96 \\
scientific\_notation.add\_coefficients\_when\_multiplying & 2700 & 836 & 30.96 \\
geometry.count\_net\_perimeter\_as\_surface\_area & 2700 & 839 & 31.07 \\
subtraction.stops\_borrow\_at\_zero & 180 & 58 & 32.22 \\
factoring.sum\_of\_squares\_factors & 2700 & 899 & 33.30 \\
subtraction.skip\_equal & 2700 & 902 & 33.41 \\
multiplication\_division.forget\_to\_add\_carried\_number & 3600 & 1277 & 35.47 \\
negative\_numbers.multiplication\_rule\_for\_addition & 2700 & 978 & 36.22 \\
subtraction.always\_borrow\_left & 180 & 67 & 37.22 \\
algebra.cancel\_across\_equals & 2700 & 1007 & 37.30 \\
order\_of\_operations.multiplication\_before\_division\_always & 2700 & 1016 & 37.63 \\
fractions.subtract\_across & 3060 & 1200 & 39.22 \\
\bottomrule
\end{tabular}
\caption{Per-malrule performance breakdown (Part 1 of 3).}
\label{tab:per_malrule}
\end{table*}

\begin{table*}[t]
\centering
\small
\begin{tabular}{llll}
\toprule
Malrule & Total & MRA & Accuracy \\
\midrule
subtraction.borrow\_no\_decrement & 180 & 71 & 39.44 \\
fractions.keep\_common\_denominator\_for\_multiplication & 3060 & 1243 & 40.62 \\
factoring.negative\_one\_factor\_forgotten & 2880 & 1189 & 41.28 \\
exponents.add\_exponents\_for\_power\_of\_power & 2880 & 1245 & 43.23 \\
negative\_numbers.negative\_swaps\_operation & 2700 & 1177 & 43.59 \\
multiplication\_division.alignment\_error\_in\_multi\_digit & 3600 & 1574 & 43.72 \\
absolute\_value.absolute\_value\_distributes & 2700 & 1188 & 44.00 \\
ratios\_proportions.additive\_instead\_of\_multiplicative & 2880 & 1306 & 45.35 \\
fractions.ignore\_denominators & 180 & 83 & 46.11 \\
algebra.distribute\_over\_non\_distributive & 2700 & 1254 & 46.44 \\
fractions.multiply\_across\_for\_division & 3240 & 1527 & 47.13 \\
statistics.ignore\_outliers\_effect & 2880 & 1388 & 48.19 \\
algebra.forget\_negative\_division & 180 & 87 & 48.33 \\
ratios\_proportions.swap\_ratios\_or\_units & 180 & 87 & 48.33 \\
ratios\_proportions.ratio\_as\_division\_only & 2880 & 1398 & 48.54 \\
scientific\_notation.ignore\_different\_powers\_of\_ten & 2700 & 1327 & 49.15 \\
measurement.inverted\_conversion\_factor & 180 & 90 & 50.00 \\
word\_problems.include\_all\_numbers\_given & 3960 & 1997 & 50.43 \\
subtraction.diff\_0\_n\_equals\_n & 2700 & 1376 & 50.96 \\
decimals.right\_align\_decimals & 2880 & 1500 & 52.08 \\
factoring.incomplete\_factoring & 2700 & 1415 & 52.41 \\
linear\_equations.slope\_is\_delta\_x\_over\_delta\_y & 2700 & 1421 & 52.63 \\
functions.function\_notation\_is\_multiplication & 180 & 95 & 52.78 \\
decimals.longer\_is\_larger & 2880 & 1540 & 53.47 \\
order\_of\_operations.pemdas\_strictly\_sequential & 2700 & 1450 & 53.70 \\
percentages.percentage\_as\_index & 2700 & 1465 & 54.26 \\
absolute\_value.inequality\_direction\_confusion & 2160 & 1178 & 54.54 \\
order\_of\_operations.strict\_left\_to\_right & 2700 & 1473 & 54.56 \\
linear\_equations.confuse\_slope\_and\_intercept\_roles & 2700 & 1508 & 55.85 \\
radicals.square\_root\_equals\_plus\_minus & 3060 & 1743 & 56.96 \\
negative\_numbers.larger\_absolute\_value\_always\_wins & 180 & 104 & 57.78 \\
decimals.whole\_number\_thinking & 2880 & 1673 & 58.09 \\
decimals.ignore\_decimal\_point & 2880 & 1718 & 59.65 \\
radicals.square\_root\_is\_divide\_by\_two & 3060 & 1842 & 60.20 \\
\bottomrule
\end{tabular}
\caption{Per-malrule performance breakdown (Part 2 of 3).}
\label{tab:per_malrule2}
\end{table*}

\begin{table*}[t]
\centering
\small
\begin{tabular}{llll}
\toprule
Malrule & Total & MRA & Accuracy \\
\midrule
rounding.leading\_zeros\_are\_significant & 3600 & 2182 & 60.61 \\
exponents.multiply\_exponents\_when\_multiplying\_powers & 2880 & 1782 & 61.88 \\
subtraction.subtract\_smaller\_from\_larger\_each\_column & 180 & 112 & 62.22 \\
fractions.denominator\_comparison\_error & 3060 & 1905 & 62.25 \\
linear\_equations.slope\_direction\_confusion & 2880 & 1826 & 63.40 \\
multiplication\_division.multiplication\_makes\_bigger & 3600 & 2299 & 63.86 \\
factoring.forget\_gcf\_first & 180 & 115 & 63.89 \\
percentages.percent\_equals\_decimal & 2700 & 1766 & 65.41 \\
exponents.forget\_exponent\_on\_coefficient & 2880 & 1914 & 66.46 \\
rounding.trailing\_zeros\_always\_significant & 3600 & 2401 & 66.69 \\
fractions.common\_denominator\_numerator & 2880 & 1943 & 67.47 \\
multiplication\_division.divide\_larger\_by\_smaller\_always & 3600 & 2471 & 68.64 \\
functions.scalar\_multiplication\_inside\_or\_outside\_same & 2700 & 1860 & 68.89 \\
percentages.add\_percentages\_directly & 180 & 124 & 68.89 \\
exponents.distribute\_exponent\_over\_addition & 180 & 133 & 73.89 \\
fractions.add\_numerators\_denominators & 180 & 134 & 74.44 \\
statistics.mode\_must\_exist & 2628 & 1985 & 75.53 \\
geometry.same\_area\_same\_perimeter & 180 & 136 & 75.56 \\
multiplication\_division.division\_makes\_smaller & 3600 & 2778 & 77.17 \\
decimals.shorter\_is\_larger & 2880 & 2262 & 78.54 \\
percentages.reverse\_percentage\_error & 2700 & 2137 & 79.15 \\
geometry.volume\_formula\_for\_surface\_area & 2700 & 2168 & 80.30 \\
linear\_equations.y\_intercept\_always\_positive & 2700 & 2170 & 80.37 \\
functions.function\_distributive\_property & 180 & 146 & 81.11 \\
radicals.distribute\_square\_root\_over\_addition & 3060 & 2547 & 83.24 \\
geometry.same\_perimeter\_same\_area & 180 & 150 & 83.33 \\
subtraction.decompose\_by\_place\_value\_label & 2700 & 2255 & 83.52 \\
exponents.negative\_exponent\_makes\_negative & 2880 & 2415 & 83.85 \\
rounding.decimal\_places\_same\_as\_sig\_figs & 180 & 153 & 85.00 \\
exponents.multiply\_base\_by\_exponent & 2340 & 2022 & 86.41 \\
radicals.negative\_outside\_same\_as\_inside & 3060 & 2691 & 87.94 \\
exponents.zero\_exponent\_equals\_zero & 2880 & 2714 & 94.24 \\
functions.same\_input\_different\_outputs\_ok & 2700 & 2628 & 97.33 \\
\bottomrule
\end{tabular}
\caption{Per-malrule performance breakdown (Part 3 of 3).}
\label{tab:per_malrule3}
\end{table*}

\subsection{Template Listing}
Tables~\ref{tab:templates_by_malrule}--\ref{tab:templates_by_malrule3} list all templates for each malrule, organized by category.

\section{Template Listing}
\label{appendix:template}

This section provides comprehensive analysis of all 498 templates across 101 malrules. The template design reflects intentional pedagogical choices grounded in learning science research, enabling rigorous cross-template generalization testing.

\subsection{Template Statistics Overview}

\begin{itemize}
    \item \textbf{Total templates}: 498 across 101 malrules (4.9 average per malrule)
    \item \textbf{Context domains}: 10 domains (63.3\% abstract, 36.7\% contextualized)---grounded in \textit{transfer of learning} research
    \item \textbf{Scaffold levels}: Basic (18.5\%), Variant (50.8\%), Context (6.2\%), Word Problem (24.5\%)---mirrors classroom instruction progression
\end{itemize}

\subsection{Context Domain Examples}

Templates embed problems in diverse real-world contexts (Table~\ref{tab:context_domains}). This design tests \textit{transfer of learning}---research shows students often fail to apply knowledge across contexts (Bransford et al., 1999). A student may correctly add fractions abstractly but fail when the same problem appears as pizza slices.

\begin{table*}[t]
\small
\centering
\begin{tabular}{p{2.5cm}p{4cm}p{8cm}}
\toprule
\textbf{Domain} & \textbf{Keywords} & \textbf{Example} \\
\midrule
Abstract (63\%) & Pure math notation & ``Calculate: $\frac{1}{2} + \frac{1}{3}$'' \\
Measurement (11\%) & Area, meters, feet & ``A rectangle has length 2.1m and width 5.4m. What is its area?'' \\
Money (9\%) & Cost, price, dollar & ``Sarah has \$0.5 and Tom has \$0.479. Who has more?'' \\
Time/Distance (4\%) & Speed, hours, miles & ``A car travels at 5.4 km/hr for 4.3 hours. Total distance?'' \\
Science (3\%) & Bacteria, wavelength & ``A wavelength of 1.13 $\mu$m is multiplied by 3.41'' \\
Sports (3\%) & Points, scores, team & ``Team A has 4x points. Team B has 1x points...'' \\
Food (3\%) & Pizza, recipe, cake & ``Sara ate $\frac{1}{4}$ of a pizza...'' \\
Temperature (2\%) & Degrees, heating & ``The temperature was $x$ degrees. It rose by 5...'' \\
Sharing (1\%) & Divided among & ``If 12 cookies are shared among 4 friends...'' \\
Elevation (1\%) & Submarine, depth & ``A submarine at -20m descends 15m more...'' \\
\bottomrule
\end{tabular}
\caption{Context domain distribution across templates.}
\label{tab:context_domains}
\end{table*}

\subsection{Scaffolded Complexity}

Each malrule follows a pedagogical progression mirroring classroom instruction (Table~\ref{tab:scaffold_levels}), grounded in Vygotsky's zone of proximal development. This scaffolding enables assessment across difficulty gradients---word problems in particular test whether misconceptions persist despite additional cognitive load from reading comprehension.

\begin{table*}[t]
\small
\centering
\begin{tabular}{p{2.5cm}p{12cm}}
\toprule
\textbf{Level} & \textbf{Characteristics} \\
\midrule
Basic (18.5\%) & Core mathematical formulation with simple values: ``Calculate: $\frac{1}{2} + \frac{1}{3}$'' \\
Variant (50.8\%) & Structural variations---larger numbers, multiple operands, negative values, edge cases: ``Calculate: $\frac{3}{4} + \frac{5}{6} + \frac{7}{8}$'' \\
Context (6.2\%) & Real-world scenario with units: ``A rope is 41.24m long and another is 2.5m. Total length?'' \\
Word Problem (24.5\%) & Full story problem requiring comprehension: ``Maria earned \$11.50 on Monday, \$8.75 on Tuesday, and spent \$4.30. How much does she have?'' \\
\bottomrule
\end{tabular}
\caption{Scaffold level distribution across templates.}
\label{tab:scaffold_levels}
\end{table*}

\subsection{Templates by Malrule}

Tables~\ref{tab:templates_by_malrule}--\ref{tab:templates_by_malrule4} list all templates for each malrule, organized by category.

\begin{table*}[t]
\small
\centering
\begin{tabular}{p{2.5cm}p{5cm}p{8cm}}
\toprule
\textbf{Category} & \textbf{Malrule} & \textbf{Templates} \\
\midrule
absolute\_value & absolute\_value\_distributes & basic\_addition, basic\_subtraction, negative\_result\_inside, multiplication\_inside, word\_problem \\
 & absolute\_value\_makes\_positive & basic\_equation, expression\_inside, inequality, compound\_expression, word\_problem \\
 & inequality\_direction\_confusion & less\_than, greater\_than \\
algebra & cancel\_across\_equals & basic\_equation, word\_problem\_context, both\_sides\_constant, three\_term\_equation, comparison\_problem \\
 & change\_side\_change\_sign & default \\
 & distribute\_over\_non\_distributive & basic\_square\_binomial, basic\_sqrt\_sum, square\_binomial\_subtraction, sqrt\_difference, word\_problem \\
 & divide\_one\_term\_only & default \\
 & forget\_negative\_division & default \\
 & variable\_letter\_has\_value & basic\_addition, basic\_subtraction, basic\_multiplication, two\_step\_equation, word\_problem \\
decimals & ignore\_decimal\_point & basic\_multiplication, division, scientific\_context, money\_context, measurement\_context, word\_problem\_context \\
 & longer\_is\_larger & basic\_comparison, ordering, money\_context, measurement\_context, number\_line, word\_problem\_context \\
 & right\_align\_decimals & basic\_addition, basic\_subtraction, money\_word\_problem, measurement\_word\_problem, three\_number\_mixed, word\_problem\_context \\
 & shorter\_is\_larger & basic\_comparison, ordering, money\_context, measurement\_context, number\_line, word\_problem\_context \\
 & whole\_number\_thinking & basic\_addition, basic\_comparison, subtraction, multiplication, word\_problem, word\_problem\_context \\
exponents & add\_exponents\_for\_power\_of\_power & basic\_power\_of\_power, with\_coefficient, numerical\_evaluation, product\_of\_powers, word\_problem, word\_problem\_context \\
 & distribute\_exponent\_over\_addition & simple\_two\_term \\
 & forget\_exponent\_on\_coefficient & basic\_power\_of\_power, larger\_coefficients, negative\_coefficient, multiple\_variables, word\_problem, word\_problem\_context \\
 & multiply\_base\_by\_exponent & simple\_numeric, larger\_exponent, word\_problem\_context \\
 & \shortstack[l]{multiply\_exponents\\\_when\_multiplying\_powers}  & basic\_two\_powers, three\_powers, numerical\_base, mixed\_operations, word\_problem, word\_problem\_context \\
 & negative\_exponent\_makes\_negative & simple\_numeric, larger\_negative\_exponent, fractional\_base, word\_problem\_scientific, word\_problem\_finance, word\_problem\_context \\
 & zero\_exponent\_equals\_zero & basic\_zero\_exponent, expression\_simplification, exponent\_rules, polynomial\_evaluation, word\_problem, word\_problem\_context \\
factoring & forget\_gcf\_first & default \\
 & incomplete\_factoring & gcf\_then\_difference\_of\_squares, gcf\_then\_trinomial, gcf\_then\_perfect\_square, nested\_factoring, word\_problem \\
 & negative\_one\_factor\_forgotten & basic\_difference\_of\_squares, coefficient\_difference\_of\_squares, trinomial\_factoring, gcf\_then\_pattern, comparison\_problem, word\_problem \\
 & sign\_errors\_in\_factoring & default \\
 & sum\_of\_squares\_factors & basic\_sum\_of\_squares, coefficient\_on\_x, both\_variables, larger\_coefficients, word\_problem \\
\bottomrule
\end{tabular}
\caption{Templates by malrule (Part 1 of 4).}
\label{tab:templates_by_malrule}
\end{table*}

\begin{table*}[t]
\small
\centering
\begin{tabular}{p{2.5cm}p{6cm}p{8cm}}
\toprule
\textbf{Category} & \textbf{Malrule} & \textbf{Templates} \\
\midrule
fractions & add\_numerators\_denominators & default \\
 & common\_denominator\_numerator & basic\_addition, word\_problem\_context, three\_fractions, word\_problem\_three\_fractions, visual\_representation, comparison\_problem \\
 & denominator\_comparison\_error & basic\_two\_fractions, three\_fractions\_ordering, word\_problem\_three\_fractions, mixed\_comparisons, word\_problem\_mixed, real\_world\_context, benchmark\_comparison \\
 & ignore\_denominators & default \\
 & keep\_common\_denominator\_for\_multiplication & basic\_common\_denominator, three\_fractions, word\_problem\_three\_fractions, mixed\_numbers, word\_problem\_mixed, larger\_denominators, word\_problem \\
 & multiply\_across\_for\_division & basic\_fraction\_division, whole\_number\_divisor, word\_problem\_whole\_divisor, whole\_number\_dividend, word\_problem\_whole\_dividend, mixed\_numbers, word\_problem\_mixed, word\_problem \\
 & multiply\_rule\_for\_addition & default \\
 & natural\_number\_bias\_numerator\_only & basic\_comparison, visual\_models, real\_world\_context, multiple\_choice, word\_problem\_multiple\_choice, ordering, word\_problem\_ordering \\
 & subtract\_across & basic\_subtraction, word\_problem\_pizza, three\_fractions, word\_problem\_three\_fractions, improper\_fractions, word\_problem\_improper, word\_problem\_measurement \\
functions & function\_distributive\_property & default \\
 & function\_notation\_is\_multiplication & default \\
 & same\_input\_different\_outputs\_ok & ordered\_pairs\_set, table\_format, graph\_points, mapping\_diagram, word\_problem \\
 & scalar\_multiplication\_inside\_or\_outside\_same & quadratic\_function, cubic\_function, absolute\_value, square\_root, word\_problem \\
geometry & count\_net\_perimeter\_as\_surface\_area & cube\_net, rectangular\_prism\_net, triangular\_prism\_net, pyramid\_net, word\_problem\_context \\
 & same\_area\_same\_perimeter & default \\
 & same\_perimeter\_same\_area & default \\
 & volume\_formula\_for\_surface\_area & rectangular\_prism, cube, larger\_dimensions, decimal\_dimensions, word\_problem \\
graphing & ignore\_coordinate\_signs & single\_point\_plotting, all\_four\_quadrants, distance\_between\_points, midpoint\_calculation, word\_problem \\
 & reverse\_coordinate\_order & default \\
linear\_equations & confuse\_slope\_and\_intercept\_roles & write\_equation, identify\_slope, identify\_y\_intercept, standard\_form\_identify, word\_problem\_context \\
 & slope\_direction\_confusion & basic\_positive\_slope, basic\_negative\_slope, larger\_magnitude, fractional\_slope, comparison\_problem, word\_problem\_context \\
 & slope\_is\_delta\_x\_over\_delta\_y & basic\_two\_points, larger\_coordinates, real\_world\_rate, negative\_coordinates, mixed\_quadrants \\
 & y\_intercept\_always\_positive & basic\_slope\_intercept, standard\_form, point\_slope\_form, two\_points\_form, word\_problem\_context \\
measurement & inverted\_conversion\_factor & default \\
 & wrong\_conversion\_factor & default \\
\bottomrule
\end{tabular}
\caption{Templates by malrule (Part 2 of 4).}
\label{tab:templates_by_malrule2}
\end{table*}

\begin{table*}[t]
\small
\centering
\begin{tabular}{p{3cm}p{5cm}p{9cm}}
\toprule
\textbf{Category} & \textbf{Malrule} & \textbf{Templates} \\
\midrule
multiplication\_division & alignment\_error\_in\_multi\_digit & basic\_2x2, word\_problem, three\_digit, money\_context, array\_model, three\_digit\_times\_three, perimeter\_then\_area, multi\_item\_purchase, tiling\_problem, calendar\_calculation \\
 & divide\_larger\_by\_smaller\_always & basic\_smaller\_dividend, fraction\_result, decimal\_friendly, word\_problem\_sharing, word\_problem\_measurement, money\_sharing, probability\_ratio, percentage\_grade, ratio\_comparison, scale\_model \\
 & division\_makes\_smaller & basic\_decimal\_division, fraction\_division, measurement\_context, sharing\_context, rate\_context, time\_conversion, capacity\_division, speed\_calculation, recipe\_scaling\_up, unit\_conversion \\
 & forget\_to\_add\_carried\_number & basic\_two\_digit\_times\_one, three\_digit\_times\_one, two\_digit\_times\_two, money\_context, word\_problem, three\_digit\_times\_two, area\_calculation, total\_cost\_bulk, distance\_calculation, array\_larger \\
 & multiplication\_makes\_bigger & basic\_decimal\_multiplication, fraction\_multiplication, money\_context, measurement\_context, percent\_discount, area\_rectangle, probability\_compound, compound\_scaling, rate\_distance, volume\_box \\
negative\_numbers & larger\_absolute\_value\_always\_wins & default \\
 & multiplication\_rule\_for\_addition & two\_negatives\_addition, positive\_minus\_negative, word\_problem\_temperature, word\_problem\_elevation, sequential\_operations \\
 & negative\_swaps\_operation & basic\_arithmetic, word\_problem\_context, multi\_step\_expression, algebraic\_context, comparison\_problem \\
 & negative\_times\_negative\_negative & basic\_multiplication, word\_problem\_temperature, word\_problem\_debt, equation\_solving, pattern\_completion, word\_problem\_multi\_step, word\_problem\_mixed\_operations, word\_problem\_distributive \\
 & two\_negatives\_always\_positive & basic\_subtraction, addition\_negatives, word\_problem\_temp, word\_problem\_money, multi\_step \\
order\_of\_operations & addition\_before\_subtraction\_always & simple\_expression, money\_transaction, temperature\_change, elevation\_change, score\_tracking \\
 & ignore\_parentheses & basic\_single\_parentheses, nested\_parentheses, multiple\_parentheses, brackets\_and\_parentheses, word\_problem \\
 & multiplication\_before\_division\_always & basic\_expression, word\_problem\_sharing, word\_problem\_measurement, word\_problem\_money, multi\_operation\_chain \\
 & pemdas\_strictly\_sequential & basic\_mult\_div, basic\_add\_sub, longer\_expressions, mixed\_operations, word\_problem \\
 & strict\_left\_to\_right & add\_mult, sub\_mult, add\_div, sub\_div, word\_problem \\
percentages & add\_percentages\_directly & default \\
 & percent\_equals\_decimal & percent\_to\_decimal, decimal\_to\_percent, percent\_in\_calculation, fraction\_to\_percent, word\_problem \\
 & percentage\_as\_index & basic\_percentage, word\_problem\_context, percentage\_increase, percentage\_decrease, comparison\_problem \\
 & reverse\_percentage\_error & basic\_reverse\_relationship, percentage\_increase, percentage\_decrease, comparison\_statements, word\_problem \\
radicals & add\_under\_common\_root & basic\_addition, basic\_subtraction, three\_radicals, word\_problem\_distance, mixed\_operations, word\_problem\_addition, word\_problem\_subtraction, word\_problem\_three\_radicals, word\_problem\_mixed \\
 & distribute\_square\_root\_over\_addition & basic\_sum\_of\_squares, pythagorean\_context, algebraic\_expression, subtraction\_variant, non\_perfect\_square, word\_problem\_subtraction, word\_problem\_algebraic \\
 & negative\_outside\_same\_as\_inside & basic\_negative\_outside, expression\_form, in\_equations, combined\_operations, word\_problem, word\_problem\_diverse, word\_problem\_combined\_ops \\
 & square\_root\_equals\_plus\_minus & basic\_square\_root, equation\_vs\_expression, negative\_radicand\_squared, expression\_simplification, word\_problem\_context, word\_problem\_non\_area, word\_problem\_equation\_vs\_expression \\
 & square\_root\_is\_divide\_by\_two & basic\_perfect\_square, larger\_perfect\_squares, area\_of\_square, equation\_solving, pythagorean\_word\_problem, word\_problem\_non\_geometric, word\_problem\_verification \\
\bottomrule
\end{tabular}
\caption{Templates by malrule (Part 3 of 4).}
\label{tab:templates_by_malrule3}
\end{table*}

\begin{table*}[t]
\small
\centering
\begin{tabular}{p{2.5cm}p{5.5cm}p{9cm}}
\toprule
\textbf{Category} & \textbf{Malrule} & \textbf{Templates} \\
\midrule
ratios\_proportions & additive\_instead\_of\_multiplicative & basic\_scaling, recipe\_scaling, speed\_problems, similar\_figures, mixture\_problems, word\_problem\_context \\
 & each\_fraction\_digit\_is\_ratio & basic\_two\_digit, three\_digit\_ratios, mismatched\_lengths, with\_zeros, word\_problem\_context \\
 & ratio\_as\_division\_only & part\_to\_part\_ratio, recipe\_scaling, paint\_mixing, distance\_rate, map\_scale, word\_problem\_context \\
 & swap\_ratios\_or\_units & default \\
rounding & decimal\_places\_same\_as\_sig\_figs & default \\
 & leading\_zeros\_are\_significant & basic\_decimal\_leading\_zeros, measurement\_context, scientific\_notation\_comparison, rounding\_to\_sig\_figs, calculation\_result, division\_word\_problem, unit\_conversion\_word\_problem, density\_calculation\_word\_problem, rate\_calculation\_word\_problem, percentage\_calculation\_word\_problem \\
 & trailing\_zeros\_always\_significant & basic\_whole\_number\_trailing\_zeros, measurement\_with\_units, scientific\_notation\_comparison, with\_without\_decimal, rounding\_application, population\_word\_problem, distance\_word\_problem, rounding\_result\_word\_problem, estimation\_word\_problem, large\_scale\_word\_problem \\
scientific\_notation & add\_coefficients\_when\_multiplying & basic\_multiplication, word\_problem\_context, multiple\_step\_calculation, compare\_results, area\_volume\_calculation \\
 & count\_all\_zeros\_for\_exponent & small\_decimal\_trailing\_zero, large\_number\_trailing\_zeros, word\_problem\_measurement, comparison\_verification, mixed\_zeros\_decimal \\
 & ignore\_different\_powers\_of\_ten & basic\_addition, basic\_subtraction, large\_exponent\_difference, negative\_exponents, word\_problem \\
 & wrong\_exponent\_sign & basic\_conversion \\
statistics & ignore\_outliers\_effect & basic\_outlier, word\_problem\_context, multiple\_outliers, symmetric\_no\_outlier, comparison\_problem, word\_problem\_expanded \\
 & mean\_without\_understanding & outlier\_high, outlier\_low, bimodal, skewed\_right, best\_measure\_question, income\_inequality\_context, real\_estate\_market, environmental\_data, daily\_routine\_outliers, word\_problem\_context \\
 & mode\_must\_exist & basic\_no\_mode, larger\_no\_mode, small\_no\_mode, has\_mode\_control, comparison\_problem, word\_problem\_context \\
subtraction & always\_borrow\_left & default \\
 & borrow\_from\_bottom & basic\_subtraction, word\_problem\_context, multi\_step\_problem, missing\_number, comparison\_problem \\
 & borrow\_no\_decrement & default \\
 & carry\_ones\_digit\_instead\_of\_tens & basic\_two\_digit, larger\_sums, three\_numbers, money\_context, word\_problem \\
 & decompose\_by\_place\_value\_label & basic\_regrouped\_tens, regrouped\_ones, multiple\_regroupings, four\_digit, word\_problem \\
 & diff\_0\_n\_equals\_n & basic\_single\_zero, multiple\_zeros, zero\_in\_ones\_place, consecutive\_zeros, word\_problem \\
 & no\_column\_limit & basic\_two\_digit, three\_digit, multiple\_carries, money\_context, word\_problem \\
 & skip\_equal & basic\_skip\_equal, multiple\_equal\_columns, word\_problem\_context, sequential\_equal\_digits, all\_equal\_digits \\
 & smaller\_from\_larger & basic\_subtraction, money\_context, measurement\_context, three\_digit, four\_digit, missing\_minuend, missing\_subtrahend, comparison, multi\_step, word\_problem\_result\_unknown, word\_problem\_missing\_minuend, word\_problem\_missing\_subtrahend, word\_problem\_comparison \\
 & stops\_borrow\_at\_zero & default \\
 & subtract\_smaller\_from\_larger\_each\_column & default \\
word\_problems & include\_all\_numbers\_given & shopping\_cart, time\_schedule, area\_calculation, collection\_combining, sharing\_division, multi\_item\_purchase, multi\_step\_time, volume\_calculation, perimeter\_with\_area, multi\_step\_collection, multi\_step\_division, mixed\_operations \\
\bottomrule
\end{tabular}
\caption{Templates by malrule (Part 4 of 4).}
\label{tab:templates_by_malrule4}
\end{table*}

\section{Framework Details}

This section provides additional details on the \textsc{MalruleLib} framework classification and template coverage.

\subsection{Classification Details}
Table~\ref{tab:strand_distribution} shows the distribution of malrules across NCTM strands.

\begin{table*}[t]
\centering
\caption{Distribution of Malrules by NCTM Strand}
\label{tab:strand_distribution}
\begin{tabular}{lcc}
\toprule
\textbf{NCTM Strand} & \textbf{Categories} & \textbf{Malrules (\%)} \\
\midrule
Number \& Operations & 4 & 54 (53.5\%) \\
Algebra & 3 & 37 (36.6\%) \\
Geometry \& Measurement & 2 & 8 (7.9\%) \\
Data \& Modeling & 1 & 4 (4.0\%) \\
\midrule
\textbf{Total} & \textbf{10} & \textbf{101 (100\%)} \\
\bottomrule
\end{tabular}
\end{table*}

\subsection{Template Coverage}
Table~\ref{tab:template_coverage} provides detailed template coverage statistics for each mathematical category.

\begin{table*}[t]
\centering
\small
\begin{tabular}{lrrrrr}
\toprule
\textbf{Category} & \textbf{Malrules} & \textbf{Templates} & \textbf{Avg} & \textbf{\% Basic} & \textbf{\% Word Prob} \\
\midrule
Multiplication \& Division & 5 & 50 & 10.0 & 8.0 & 8.0 \\
Subtraction & 11 & 47 & 4.3 & 23.4 & 21.3 \\
Fractions & 9 & 45 & 5.0 & 20.0 & 37.8 \\
Radicals & 5 & 37 & 7.4 & 16.2 & 43.2 \\
Exponents & 7 & 34 & 4.9 & 11.8 & 35.3 \\
Decimals & 5 & 30 & 6.0 & 23.3 & 26.7 \\
Order of Operations & 5 & 25 & 5.0 & 16.0 & 24.0 \\
Negative Numbers & 5 & 24 & 4.8 & 16.7 & 41.7 \\
Statistics & 3 & 22 & 7.3 & 9.1 & 18.2 \\
Linear Equations & 4 & 21 & 5.2 & 19.0 & 14.3 \\
Rounding & 3 & 21 & 7.0 & 14.3 & 47.6 \\
Algebra & 6 & 18 & 3.0 & 50.0 & 16.7 \\
Factoring & 5 & 18 & 3.6 & 22.2 & 16.7 \\
Ratios \& Proportions & 4 & 18 & 4.5 & 16.7 & 16.7 \\
Percentages & 4 & 16 & 4.0 & 18.8 & 18.8 \\
Scientific Notation & 4 & 16 & 4.0 & 25.0 & 18.8 \\
Absolute Value & 3 & 12 & 4.0 & 25.0 & 16.7 \\
Functions & 4 & 12 & 3.0 & 16.7 & 16.7 \\
Geometry & 4 & 12 & 3.0 & 16.7 & 16.7 \\
Word Problems & 1 & 12 & 12.0 & 0.0 & 0.0 \\
Graphing & 2 & 6 & 3.0 & 16.7 & 16.7 \\
Measurement & 2 & 2 & 1.0 & 100.0 & 0.0 \\
\midrule
\textbf{Total} & \textbf{101} & \textbf{498} & \textbf{4.9} & \textbf{18.3} & \textbf{24.5} \\
\bottomrule
\end{tabular}
\caption{Template coverage across 22 mathematical categories. Each malrule has an average of 4.9 templates (ranging from 1 to 13), enabling diverse problem generation for each misconception. Template types include basic formulations (18.3\%), real-world contexts (10.0\%), word problems (24.5\%), and structural variants (47.2\%). The 498 templates span 334 unique template patterns, providing rich variety for cross-template generalization testing.}
\label{tab:template_coverage}
\end{table*}

\end{document}